%% file: cas-dc-template.tex
\documentclass[a4paper,fleqn]{cas-dc}

\usepackage[authoryear,longnamesfirst]{natbib}
\usepackage{adjustbox}

\usepackage{multirow}
\usepackage{tabulary}
\usepackage{colortbl}
\usepackage{booktabs}
\usepackage{algorithm}
\usepackage{algorithmic}

\newcommand{\etal}{~\textit{et al.}}

\newcommand{\eg}{\textit{e.g.}}

\def\tsc#1{\csdef{#1}{\textsc{\lowercase{#1}}\xspace}}
\tsc{WGM}
\tsc{QE}
\tsc{EP}
\tsc{PMS}
\tsc{BEC}
\tsc{DE}

\begin{document}

\shorttitle{Few-Shot Segmentation with Global and Local Contrastive Learning}

\shortauthors{Weide Liu~\etal}

\title [mode = title]{Few-Shot Segmentation with Global and Local Contrastive Learning}

\author[label1]{Weide Liu}%% Author name
\author[label1]{Zhonghua Wu}
\author[label2]{Henghui Ding}
\author[label3]{Fayao Liu}
\author[label3]{Jie Lin}
\author[label1]{Guosheng Lin}
\author[label4]{Wei Zhou}

\affiliation[label1]{organization={School of Computer Science and Engineering, Nanyang Technological University}, Singapore}
\affiliation[label2]{organization={Institute of Big Data, Fudan University}, China}
\affiliation[label3]{organization={Institute for Infocomm Research (I$^2$R)-Agency for Science, Technology and Research}, Singapore}
\affiliation[label4]{organization={Cardiff University}, UK}

\input{0_abstract}

\maketitle

\input{1_Introduction}
\input{2_related_work}
\input{3_task_defination}
\input{4_method}

\input{5_experinment}
\input{6_conclusion}

\bibliographystyle{cas-model2-names}
\bibliography{cas-refs}

\end{document}

%% file: 0_abstract.tex
\begin{abstract}
This work addresses the challenging task of few-shot segmentation. Previous few-shot segmentation methods mainly employ the information of support images as guidance for query image segmentation. Although some works propose to build a cross-reference between support and query images, their extraction of query information still depends on the support images. In this paper, we propose to extract the information from the query itself independently to benefit the few-shot segmentation task. To this end, we first propose a prior extractor to learn the query information from the unlabeled images with our proposed global-local contrastive learning. Then, we extract a set of predetermined priors via this prior extractor. With the obtained priors, we generate the prior region maps for query images, which locate the objects, as guidance to perform cross-interaction with support features. In such a way, the extraction of query information is detached from the support branch, overcoming the limitation by support, and could obtain more informative query clues to achieve better interaction. Without bells and whistles, the proposed approach achieves new state-of-the-art performance for the few-shot segmentation task on public datasets.
\end{abstract}

%% file: 1_Introduction.tex
\section{Introduction}
\input{Figures/first}
Semantic segmentation is a fundamental task that involves classifying each pixel into a particular class. Deep learning has achieved remarkable success in fully supervised semantic segmentation~\cite{long2015fully}. However, this approach has some intrinsic limitations, such as requiring a large number of pixel-level annotated image samples for model training and abundant annotations for novel classes when extending the current segmentation model to new classes.

Few-shot segmentation has been proposed to address these issues by training a network to predict segmentation masks for novel classes with only a few annotated novel class training samples. Currently, state-of-the-art few-shot segmentation methods~\cite{zhang2019canet,wang2019panet,ppnet} utilize support images as guidance for query image segmentation with a two-branch architecture. However, they only use support images' information as guidance for query mask prediction and do not consider clues from the query images themselves. To use the query information, CRNet~\cite{crnet} proposes a cross-reference mechanism that enables interaction between support and query image features. Similarly, PANet~\cite{wang2019panet} uses prototype alignment regularization to align the query and support prototypes. Although these methods build cross-guidance between query and support images, they are limited to labeled support images since all information extraction/propagation, such as support-to-query and query-to-support, essentially rely on support masks.

In this work, we argue that detaching query information extraction from the support branch and generating query clues independently can enhance few-shot segmentation performance and generalization. To this end, we propose a prior extractor to learn query information from the unlabeled query images themselves with self-supervised learning.

Contrastive learning is among the most promising directions in self-supervised learning methods~\cite{contrastivelearn}. The process involves transforming images into different variants, using contrastive loss to minimize the feature distances between the variants from the same images, and maximizing the feature distances obtained from different images. The objective of contrastive learning is to learn a predetermined prior of the objects without labeled data, which can bridge the gap between fully and less supervised classification. However, the current state-of-the-art design~\cite{moco} is sub-optimal for segmentation tasks for two reasons. Firstly, image segmentation is a pixel-level classification task where both local and global representations are crucial, but \cite{moco} is only designed for global image-level representation. Secondly, there are often multiple objects co-existing in one image, such as a keyboard, desk, and computer. However, the global contrastive loss cannot distinguish between these different objects within the same image.

To address these issues, we propose a predetermined prior learning method to obtain more distinguishable image features for the few-shot semantic segmentation task. We leverage both global and local contrastive losses to learn a prior extractor for few-shot image segmentation. A global contrastive loss is applied to the global representations to minimize the feature distances obtained from different variants of an identical image. To further differentiate different objects within the same image, we obtain local representations by dividing the image into local patches, with each patch containing similar features and contexts. We then apply a local contrastive loss between the local patches to learn a local predetermined prior. Similar to the global contrastive loss, the local contrastive loss aims to maximize the feature distance between different patches, such as computer patches and keyboard patches. We use both global and local contrastive losses to train our prior extractor to leverage both advantages.

We utilized the previously mentioned self-supervised learning method to train a prior extractor on the available unlabeled data, which holds the predetermined prior for the query images. Subsequently, we developed a new few-shot segmentation architecture, as illustrated in Figure~\ref{Fig:introduction}, to extract semantic segmentation information of the query category from the priors. In this approach, the given query images undergo encoding using the prior and feature extractors, resulting in prior features and bridge features, respectively. The target objects are then identified by computing pixel-wise similarity maps between the prior and bridge features. This method enables us to obtain the prior region maps of the query images utilizing their own information, thereby addressing the aforementioned limitations. Furthermore, the support images undergo encoding by the same feature extractor to be projected onto the same feature space as the bridge features. Subsequently, we establish cross-interactions between the support and bridge features to enhance the segmentation performance.

Our main contributions can be summarized as follows:
\begin{itemize}

\item To the best of our knowledge, we are the first to utilize self-supervised feature learning methods on unlabeled query images to benefit the few-shot segmentation task. Specifically, we propose a prior extractor to generate maps of prior regions from the query images themselves to guide the final query mask prediction.

\item We propose a global and local contrastive loss to train the prior extractor, making contrastive learning more suitable for few-shot segmentation tasks. With our proposed global-local learning, the query branch independently extracts informative clues from the query images themselves, which greatly enhances the cross-interaction between the query and support.

\item Our network achieves state-of-the-art results on the COCO datasets.
\end{itemize}

%% file: Figures/first.tex
\begin{figure}[t]
\centering
    \includegraphics[width=1\linewidth]{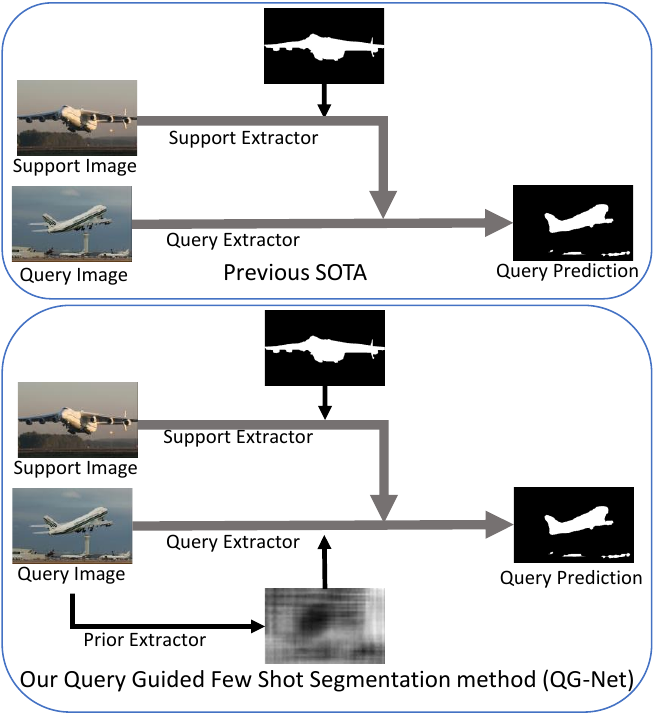}
    \caption{Comparison between the pipeline of our proposed Query Guided Network with previous state-of-the-art (SOTA) few-shot segmentation methods. Previous works (upper part) only employ support images' information as guidance for query mask estimation, while our QGNet (lower part) utilizes the clues of query images with query extractor as guidance for final query mask prediction.}
    \label{Fig:introduction}
\end{figure}

%% file: 2_related_work.tex
\section{Related Work}
\input{Figures/overall}
\subsection{Fully Supervised Semantic Segmentation}
Semantic segmentation is a crucial computer vision task that involves predicting a predefined class for each pixel in an image. Recent state-of-the-art methods rely heavily on fully convolutional networks (FCN)\cite{long2015fully}. The encoder-decoder\cite{chen2018deeplab} architecture is a popular choice for generating high-resolution prediction maps for semantic segmentation. The encoder gradually downsamples the feature maps to obtain a large field-of-view, while the decoder recovers the fine-grained information. To avoid losing the resolution of the feature maps while keeping the field-of-view large, dilated convolution~\cite{chen2018deeplab} increases the field-of-view without decreasing the feature map size or increasing the number of parameters. In our network, we follow the encoder-decoder structure with dilated convolution to produce the query segmentation masks.

\subsection{Few-shot Segmentation}
Early approaches for few-shot segmentation adopt a two-branch architecture to condition on a given support example. For instance, Shaban et al.~\cite{shaban2017one} introduced the first such model (OSLSM) with a conditioning branch that generates parameters for a query segmentation network. Subsequent methods like co-FCN~\cite{rakelly2018conditional} and CANet~\cite{zhang2019canet} improved this paradigm by embedding support features into the query branch (via feature concatenation and attention) and by iteratively refining the predicted masks. Another prominent direction is to use metric learning with class prototypes. Dong and Xing~\cite{Dong2018FewShotSS} first proposed to compute a prototype representation from the support mask and match query features to this embedding for segmentation, an idea also exploited by the similarity guidance network SG-One~\cite{zhang2018sg}. 

In terms of enhanced feature matching, HDMNet~\cite{Peng_2023_CVPR} introduce hierarchical dense correlation distillation using transformer-based multi-scale matching to mitigate overfitting. Xu et al.~\cite{Xu_2023_ICCV} propose SCCAN, employing self-calibrated cross-attention to better align query-support patches, and further improve segmentation accuracy via their ambiguity elimination strategy (AENet)~\cite{Xu_2024_ECCV}. Moon et al.~\cite{Moon_2023_ICCV} present MSI, maximizing support-set information by enriching correlation maps to handle small or ambiguous query targets.

Another promising trend involves leveraging external knowledge and pretrained models. MIANet~\cite{Yang_2023_CVPR} aggregate instance-specific and general class prototype embeddings guided by textual semantics to overcome class bias. Zhu et al.~\cite{Zhu_2024_CVPR} propose LLaFS, uniquely utilizing large language models (LLMs) to guide segmentation with language-driven region attributes, demonstrating substantial improvements in few-shot generalization. Meanwhile, visual prompting techniques have also emerged, such as the multi-scale visual prompts introduced by Hossain et al.~\cite{Hossain_2024_CVPR}, designed for generalized few-shot segmentation tasks without altering transformer weights.

The two-branch network architecture is a common approach in few-shot segmentation tasks. Various prior studies \cite{ppnet,pfenet,crnet,liu2022crcnet,liu2022few,liu2024harmonizing,fss1000,rakelly2018conditional,siam2019adaptive,zhang2018sg,amp,mdl,pmm} have employed methods to encode category features from labeled support images for guiding query mask prediction. However, these methods primarily rely on support images for guidance and do not fully exploit information from query images.
In pursuit of better leveraging query image information, CRNet~\cite{crnet} introduces a cross-reference mechanism to facilitate interaction between query and support images, enhancing model training. PANet~\cite{wang2019panet} incorporates a prototype alignment network to guide segmentation for both query and support images. SSNet~\cite{bmvc_few_shot} introduces a self-supervised approach, enhancing performance by introducing random pseudo-classes in the background of query images through superpixel segmentation.
PFENet~\cite{pfenet} employs a prior-guided feature enhancement network for comparing query and support features through pyramid feature comparison.
Building upon these advancements, methods like ASGNet~\cite{asgnet}, RePRI~\cite{repri}, CWT~\cite{cwt}, and ABPNet~\cite{dong2021abpnet} have aimed to mitigate model bias and adapt more effectively to novel classes. ASGNet dynamically determines the number of prototypes and their spatial dimensions. RePRI fine-tunes models on support images to enhance adaptation to novel classes. In contrast, CWT adjusts classifier weights using episodic training within a self-attention block, and ABPNet employs an attention mechanism with a meta-training strategy to predict task-specific backgrounds.
Additionally, SCL~\cite{scl} introduces self-guided learning to recover lost critical information during the encoding process. This method employs masked Global Average Pooling (GAP) to encode both covered and uncovered foreground regions, leading to improved segmentation performance on query images.
Nevertheless, despite these methods aligning both support and query clues and guiding each other, the query mask prediction is still limited to the labeled support images as their information extraction of query images still depends on support. In this work, we contend that training an image prior extractor using unlabeled query images could enhance the query segmentation mask prediction.

\subsection{Contrastive learning}
Self-supervised contrastive learning methods, such as MoCo~\cite{moco,mocov2} and Bootstrap~\cite{Bootstrap}, have gained popularity in computer vision for learning a feature extractor from unlabeled images as pre-training for downstream tasks. These methods aim to pull representations of different views of the same image closer together while pushing away representations of different images. SimCLR~\cite{simclr} proposes a straightforward framework for self-supervised learning by applying contrastive learning to representations of the same image with different data augmentations. MoCo~\cite{moco,mocov2} increases the negative memory bank size with a moving average network, known as a momentum encoder. 
RegionContrast~\cite{hu2021region} improves similarity between semantically similar pixels while maintaining discrimination. It leverages a memory bank and region centers for efficient feature storage, enabling region-level contrastive learning, which is more memory-efficient than pixel-level methods.
IIC~\cite{iic} focuses on maximizing mutual information between class assignments in each pair, a method designed for unsupervised image classification and segmentation.
Achanta~\cite{achanta2008salient} propose to identify salient regions in images by utilizing low-level features related to brightness (luminance) and color, while Krishna et al.~\cite{chaitanya2020contrastive} propose strategies to enhance semi-supervised segmentation of volumetric medical images, incorporating domain-specific and problem-specific cues. These strategies lead to improved performance, as demonstrated on three MRI datasets with limited annotations, surpassing other self-supervised and semi-supervised methods.
DetCo~\cite{xie2021detco} introduces multi-level supervision for intermediate representations and employs contrastive learning between the entire image and local patches. This design ensures consistent and discriminative global and local representations across feature pyramid levels, benefiting both detection and classification.

In this work, we leverage contrastive learning to train a prior extractor from the unlabeled query images to enhance few-shot segmentation. By using contrastive loss, our prior extractor can capture informative features from unlabeled query images to support query segmentation without relying solely on labeled support images.

\input{Figures/moco_vs_ours}

%% file: Figures/overall.tex
\begin{figure*}[t]
\centering
    \includegraphics[width=1\linewidth]{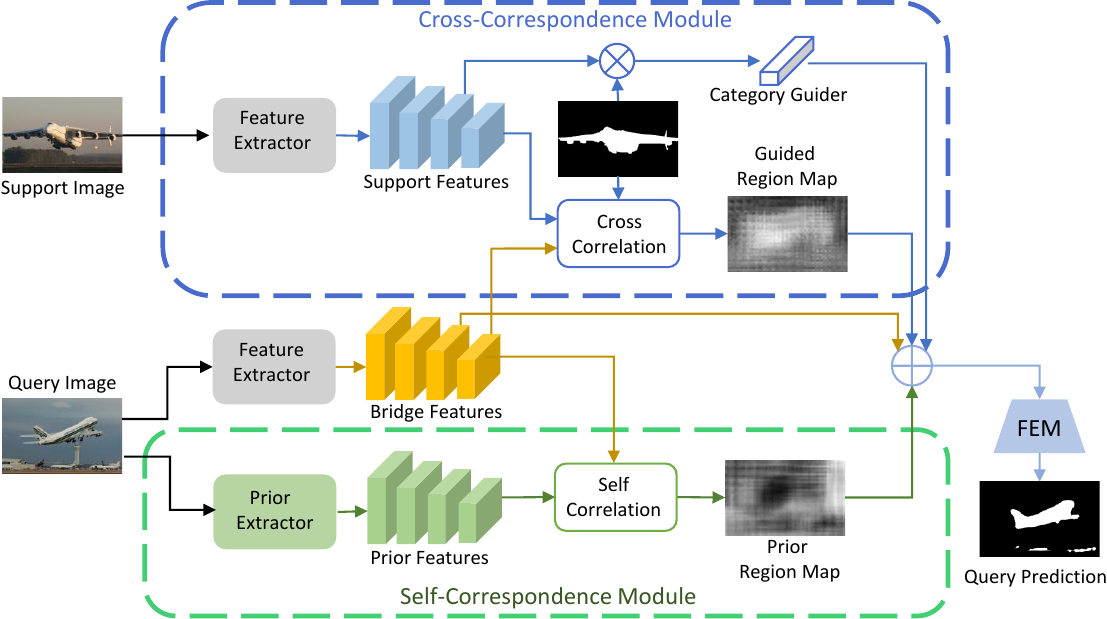}
    \caption{Our proposed method consists of a self-correspondence module and a cross-correspondence module. Unlike the previous SOTA, a self-correspondence module (green) is proposed to extract prior features and generate a prior region map to locate the target object regions from the query image itself. A Cross-correspondence module (blue) is proposed to generate a guided region map to identify the query object region with the category guide from the masked support feature. Finally, the prior region map and guided region map are concatenated with category and bridge features for the final query mask prediction (FEM). }
    \vspace{+0.36cm}
    \label{fig:overall}
        % \vspace{-0.6cm}
\end{figure*}

%% file: Figures/moco_vs_ours.tex
\begin{figure*}[t]
\centering
    \includegraphics[width=0.85\linewidth]{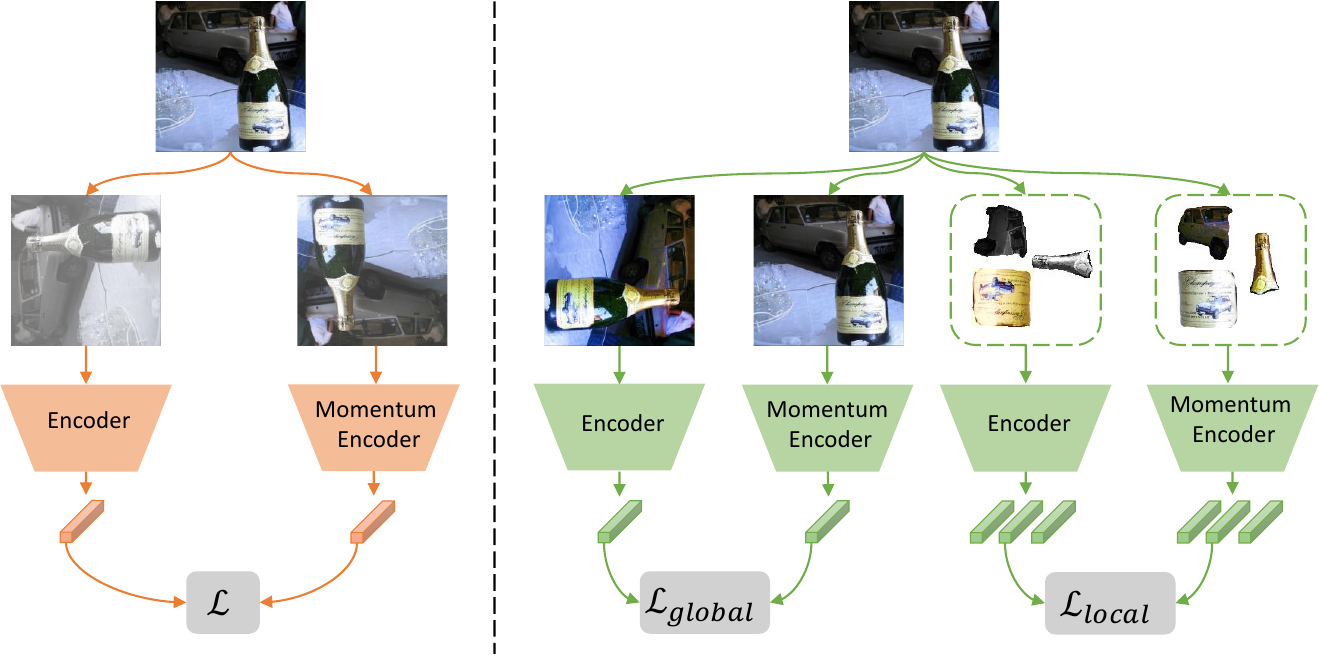}
    \caption{The difference between conventional contrastive learning and our proposed global-local contrastive learning. Conventional contrastive learning methods (left) only learn contrast from a global perspective. Our global-local contrastive learning learns the contrast with two additional local patches as input and builds a contrastive loss across the global and local representation.}
    \label{moco_vs_ours}
\vspace{+0.36cm}
\end{figure*}

%% file: 3_task_defination.tex
\section{Task Definition}
Few-shot semantic segmentation aims to perform segmentation with only a few annotated examples on novel classes. We divide the images into two sets, the target images are called query set $\mathcal{Q}$ and the annotated images serve as support set $\mathcal{S}$. Provided $K$ annotated samples from the support set, the model aims to predict binary masks for the query images, while the annotated support images should provide the foreground categories. 
The categories in the training image set have no overlap with the test image set. 

Given a network $\mathcal{R}_{\theta}$ parameterized by ${\theta}$, in each episode, we first sample a target category $c$ from the dataset $\mathcal{C}$. Based on the sampled classes, we then collect $K+1$ labeled images $\{(x_s^1,y_s^1),(x_s^2,y_s^2),...(x_s^k,y_s^k),(x_q,y_q) \}$ from the dataset, which contains at least one object belongs to the sampled category $c$. Among them, we sample $K$ labeled images constitute the support set $\mathcal{S}$, and the last one serves as the query set $\mathcal{Q}$. After that, we make predictions on the query images by inputting the support set and the query image into the model $\hat y_{q}=\mathcal{R}_{\theta}(\mathcal{S},x_q)$. At training time, we learn the model parameters $\theta$ by optimizing the cross-entropy loss $\mathcal{L}(\hat y_{q},y_{q})$, and repeat such procedures until convergence.

%% file: 4_method.tex
\section{Method}
Few-shot semantic segmentation task aims to train a network to predict novel classes with only a few annotated data. Previous works~\cite{crnet, zhang2019canet, pfenet, ppnet, pmm} solve this task by learning a network with annotated support images. However, we aim to extract information from query images themselves to improve the prediction of query masks. Our proposed method comprises two main modules: the self-correspondence module and the cross-correspondence module, as shown in Figure~\ref{fig:overall}.
In the self-correspondence module, we train a prior extractor from unlabeled query images using global and local contrastive losses. We store prior features obtained with this method to propagate stored prior information to bridge features, obtaining object regions as prior region maps. The cross-correspondence module generates guided region maps and category guider to determine query category information with annotated support images, allowing for the final prediction of query masks.

This section introduces the global contrastive loss to differentiate dissimilar image features and the proposed local contrastive loss to differentiate different object patches. The quality of patches is crucial for local contrastive learning, which impacts prior feature generation. We analyze different ways of generating patches to calculate the local contrastive loss and optimize the prior extractor.
We then describe how the prior region map and guided region map are generated from our trained prior extractor and feature extractor. Finally, we present an interesting discovery: the prior region maps have low correspondence values at object regions, while conventional guided region maps have high correspondence values at object regions.

\subsection{Prior Extractor}
\subsubsection{Global Contrastive Learning}
The global contrastive loss~\cite{contrastivelearn} has been widely used for pre-training by distinguishing features obtained from different images. In this paper, we also use the global contrastive loss to learn a global level prior to distinguishing the image level features. 

Following~\cite{mocov2}, we random sample one image from the dataset and transform the image into two different variants as a positive pair. For the negative pairs $I_k$, we sample one image from the queue. Consider we encode one query image $I_q$ to vector $q$, and encode a set of reference images as keys to vectors $k_{0,1,2...}$. 
We use the InfoNCE~\cite{infoace} as our contrastive loss:

\begin{equation}
\small
\mathcal{L}_{global} = -\log \frac{\exp(q{\cdot}k_0 / \tau)}{\sum_{i=1}^{K}\exp(q{\cdot}k_i  / \tau) + exp(q{\cdot}k_0 / \tau)} . 
\label{eq:infonce}
\end{equation}
In line with previous approaches~\cite{moco,mocov2}, we adopt $K=65536$ negative samples for our method. $k_0$ denotes the positive pair from the same image, and we set the temperature hyper-parameter $\tau$~\cite{tau}. The contrastive loss's value is low when the feature similarity is high for positive pairs and low for negative pairs.

To maintain a large dictionary size, we build a dynamic dictionary following MoCo~\cite{moco}. The keys of the dictionary are updated by replacing the oldest mini-batch with the new one in a momentum-based manner.

\subsubsection{Local Contrastive Learning}

The global contrastive loss aims to learn a global-prior feature by teaching the model to distinguish dissimilar images. However, several limitations still exist in applying global contrastive loss for segmentation task pre-training, 
1) The global contrastive loss only discriminates between dissimilar images and does not consider different objects within a single image.
2) Instances that commonly coexist might have a negative effect with only global contrastive loss. For instance, the keyboard and computer frequently appear together in the same image, but they typically need to be segmented into different classes.

To address these limitations, we propose a local contrastive loss to differentiate between different objects and improve the few-shot segmentation task. We divide the images into local patches and pre-train the model using the local contrastive loss on these patches. After introducing the local contrastive loss, we will explain the methods used to generate the patches and demonstrate how they can be used to improve few-shot segmentation tasks.

\textbf{Local contrastive loss.}
We aim to distinguish the patches by applying the local contrastive loss on the patches generated with the above methods. Specifically, we randomly sample one query patch and encode it to $q_p$, and a set of reference patches as our reference patches key, $k_{p0,p1,...}$. A local contrastive loss aims to minimize the distance between similar patches and maximize the distance between dissimilar patches. Thus, our local contrastive loss can be formulated as:
\begin{equation}
\small
\mathcal{L}_{local} = -\log \frac{\exp(q_p{\cdot}k_{p0} / \tau)}{\sum_{i=1}^{M}\exp(q_p{\cdot}k_{pi} / \tau) + exp(q_p{\cdot}k_{p0}/ \tau) } . 
\label{eq:local contrastive}
\end{equation}
Here, we keep $M=65536$ negative sample patches in the queue, and the $k_{p0}$ is the positive patch pairs with different views from the same patch, the $k_{pi}$ denotes the negative patch pairs, and the $\tau$ is a temperature hyper-parameter~\cite{tau}. 

We use both local and global contrastive loss to generate our query prior:
\begin{equation}
\small
\mathcal{L} = \mathcal{L}_{local} + \mathcal{L}_{global}.
\label{eq:fused contrastive}
\end{equation}

\textbf{Patch generation method 1: Felzenszwalb’s efficient graph based segmentation.}
To guarantee that the generated patches align with the same semantic categories, follow~\cite{felzenszwalb2004efficient}, we segment the images by assigning the local patches with the same contexts. As shown in algorithm~\ref{alg:A}, at first, we build an undirected graph $G=(V, E)$, the initial elements $v_i$ in $V$ are pixels, the edges $E$ are the edges connecting the pixels.  The weight $w$ of the edges is the dissimilarity of two pixels connected by the edge. We define the weight of edge $e_{i,j}$ with their pixel intensity $w_{i,j} = I_i - I_j$. In the beginning, we treat each pixel as a component and merge the components connected with low weight as one component until all the edges have been calculated. 
MInt($C_i$, $C_j$) is the internal difference in the components $C_i$ and $C_j$. Each final component serves as a local patch for our prior generation.  

\begin{algorithm}
\caption{Generate local patches with Felzenszwalb’s efficient graph based segmentation~\cite{felzenszwalb2004efficient} }
\label{alg:A}
\begin{algorithmic}
\STATE Input: One image
\STATE Output: $S=(C_1,C_2,...,C_r)$
\STATE Initialization: We build an undirected graph based on the image color $G=(V, E)$ with $n$ vertices and $m$ edges, where $v_i$ $\in$ $V$, $V$ denotes the image pixels, $e$ $\in$ $E$, $E$ denotes the correspondences of the neighboring vertices, and $w$ denotes the weight of $e$.
\STATE Sort the edge $E$ with their edge weight.
\STATE $S^0$: Where each $v_i$ is each pixel. 
\FOR{$q = 1,2,..,m$}
\STATE Construct $S^q$ based on $S^{q-1}$
\STATE Let $o_q = (v_i,v_j)$
\STATE Let $v_i \in C_i$ and $v_j \in C_j$
\IF{$C^{q-1}_{i} \ne C^{q-1}_{j}$ and $w(o_q) \leq MInt(C^{q-1}_{i},C^{q-1}_{j}) $} 
\STATE $S^q = S^{q-1} \cup C^{q-1}_{i} \cup C^{q-1}_{j})$ 
\ELSE 
\STATE $S^q = S^{q-1}$
\ENDIF
\ENDFOR
% \RETURN Each component
\end{algorithmic}
\end{algorithm}

\textbf{Patch generation method 2: Simple Linear Iterative Clustering (SLIC).}
We generate the image patches by clustering pixels based on their color similarity and proximity in the image plane. Following~\cite {slic}, we fuse both the color and position information of each pixel into a five-dimension format ($labxy$). Here $lab$ denotes the color information and $xy$ denotes the spatial position. To cluster the 5D spaces pixels, new distance measures between pixels need to be used. 
As shown in Algorithm~\ref{alg:B}, for every input image, we choose $K$ cluster centers $C_k$ with regular grid intervals $S=sqrt(N/K)$, where $N$ denotes the number of pixels of the input image. We assume that the cluster center connects to the associated pixels within a $2S\times2S$. We calculate the 5D distance ($D_{s}$) between the pixels $k,i$ with the following equation:
\begin{equation}
    D_s = D_{lab} + D_{xy}/S\times m. 
    \label{equ:ds}
\end{equation}
In the equation~\ref{equ:ds}, $D_{lab}$ denotes the color distance and $D_{xy}$ denotes the spatial distance:
\begin{equation}
    D_{lab} = \sqrt{((l_k-l_i)^2+ (a_k-a_i)^2+ (b_k-b_i)^2)} ,
\end{equation}

\begin{equation}
    D_{xy} = \sqrt{((x_k-x_i)^2+ (y_k-y_i)^2)}.
\end{equation}
In equation~\ref{equ:ds}, we normalize the spatial distance by dividing the interval $S$ and the variable $m$ aims to control the compactness of the superpixel. 

We re-center the cluster center to the lowest gradient position within $n\times n$ distance to the existing center (we choose $n=3$) for every cluster center. The gradient generated with : 
\begin{equation}
\begin{aligned}
    G_{xy} = ||I(x+ 1,y)-I(x-1,y)||^2+ \\
    ||I(x,y+ 1)-I(x,y-1)||^2.
    \label{equ: Gradient}
\end{aligned}
\end{equation}
Where $x,y$ denotes the pixel positional and the $I_{x,y}$ denotes the corresponding color feature vector—each pixel connected to its nearest cluster center within the search distance. 
Furthermore, a new center is generated by averaging all the pixels within the same cluster. 
The process of associating pixels with the nearest cluster center and re-centering the cluster is repeated until convergence. 
When there remain unconnected pixels $E$ less than the $threshold$, we enforce them to connect to the largest neighboring cluster and stop the process of clustering.
Each superpixel (cluster) serves as a local patch for our prior generation.  

\begin{algorithm}
\caption{Generate local patches with Simple Linear Iterative Clustering}
\label{alg:B}
\begin{algorithmic}
\STATE {Initialize cluster centers $C_k=[C_1,C_2,C_3,...]$ with a grid distance S} 
\STATE Re-center the cluster within $3\times 3$ regions to the lowest gradient position with equation~\ref{equ: Gradient}
\REPEAT 
\FOR{Each cluster}
\STATE Assign the pixels to the cluster with the distance measure in equation~\ref{equ:ds}
% \STATE Compute new cluster centers and residual error E.
\ENDFOR
\STATE Compute new cluster centers and residual error $E$.
\UNTIL{$E <= threshold$}
\RETURN Each cluster %serves as a local patch.  
\end{algorithmic}
\end{algorithm}

\subsection{Generate region maps for few-shot segmentation tasks}
As shown in Figure~\ref{fig:overall}, in this section, we show how to generate the prior region map and guided region map from our pretrained prior extractor and the feature extractor to benefit the few-shot segmentation tasks. 

We set $I_q$ denotes the query images, $I_s$ denotes the support images, and $M_s$ denotes the support annotations. $F_p$ denotes our pretrained prior extractor, and $F$ denotes the feature extractor. We encode the images with:
\begin{equation}
    X_q = F(I_q), X_s = F(I_s) ,P_q = F_p(I_q).
\end{equation}

\subsubsection{Generate prior region maps} \label{sec: generage_self_cores}
For every query image, we retrieve the query information from our prior features by calculating the correlation between the bridge features $X_q$ and $P_q$.
Specifically, we generate the prior region maps with the following steps: we first calculate the pixel-wise cosine similarity $cos(x_q, p_{q'})$ between the bridge feature $x_q \in X_q$, and the prior feature $p_{q'} \in P_q$:
\begin{equation}
\label{eq:cosine}
cos(x_q, p_{q'}) = \frac{x_q^T p_{q'}}{\left \| x_q \right \| \left \| p_{q'} \right \|}\quad .
\end{equation}
where $q,q' \in \{1,2,3,...,hw\}$ and $h,w$ denote the feature size.
For the prior region map, we locate the object by defining the confident background with the maximum similarity value alone the all prior features and combine them as the similarity map:

\begin{eqnarray}
\label{eqn:maxcosine}
s_{q} &=& \max\limits_{{q'}\in \{1, 2, ..., hw\}}(cos(x_q, p_{q'})),\\
\label{eqn:prior}
S_Q &=& [s_1, s_2, ..., s_{hw}] . %\in \mathbb{R}^{hw\times 1}.
\end{eqnarray}
We normalize the correspondence value to $(0,1)$ with: 
\begin{equation}
\label{eqn:norm}
S_Q = \frac{S_Q - \min(S_Q)}{\max(S_Q) - \min(S_Q) + \epsilon}.
\end{equation}
where $\epsilon$ is set to $1e-7$ in our experiments.
Finally, we reshape $S_Q$ into $h\times w\times 1$ to the same size as our final prior region map to locate the objects. To be noticed, as shown in Figure~\ref{fig:overall}, the prior features are obtained directly from our global-local contrastive learning using the query image. These features are extracted without any labeled information, relying solely on unsupervised learning. \newline
In contrast, the bridge features are derived from the query image, yet they are generated using a shared Feature Extractor with the support image. The parameters of this Feature Extractor have been fine-tuned using labeled training images, which provide domain knowledge.

\subsubsection{Generate guided region maps}
Following the same step as~\ref{sec: generage_self_cores}, we generate the guided region maps with the support features $X_s$ and bridge features $X_q$ within the cross-correspondence module. We filter out the irrelevant support features by multiplying the support features with the annotated support mask $M_s \times X_s$. In this way, the bridge feature pixels yield no correspondence with the background and only correlate with the target object area. Confident query object regions are then identified by selecting the highest cosine similarity for each bridge feature across all filtered support features.

Subsequently, we combine the prior and guided region maps to produce our final correspondence maps, guiding our network's predictions. Additionally, to directly provide category information without unrelated information, we derive the category guider by multiplying the support features with the annotated support mask. We then concatenate the bridge features, the category guider, the prior region maps, and the guided region maps into a final feature group, which is processed by FEM~\cite{pfenet} as a decoder to generate our final query masks.

\subsection{Opposite Region Maps}
As shown in Figure~\ref{fig:visulization}, we discover that the prior region map generated by bridge features and prior features has a low correspondence value at the object regions while having a high value at the background regions. Our conjecture is that the prior learns the object features from the unlabeled images and stores the information within the prior. The learned object features have been transferred to a new feature space, which is dissimilar to the features generated from the bridge encoder. However, the irrelevant objects (\eg, the background) remain in a similar feature space, which indicates high similarity to the features from the bridge encoder. So the prior region maps locate the target objects with the low similarity value. On the other hand, the support feature and bridge feature stay in the same feature space and are generated by different images; a high similarity score indicates high confidence that the regions contain similar objects.

%% file: 5_experinment.tex
\input{Tables/sota_voc}
\section{Experiment}
\input{Figures/visualize_slic}
\input{Figures/visualize}

\input{Tables/fbiou-voc}

\input{Tables/sota_coco}

\input{Tables/fbiou-coco}

\input{Tables/ablation}
\input{Tables/ablation_global_local}
\input{Tables/ablation_iou}
\subsection{Implementation details}
We trained the prior extractor using SGD optimizer with InfoNCE loss~\cite{infoace} and followed MOCO-v2~\cite{mocov2} to apply data augmentation techniques such as blur augmentation, color distortion, random crop, and random flip. The entire dataset was used to train the prior extractor for the COCO dataset with 200 epochs and for the PASCAL VOC dataset with 1000 epochs.

To ensure a fair comparison with previous methods~\cite{pfenet, crnet, zhang2019canet, wang2019panet, ppnet}, we selected multiple backbones, including Resnet101, Resnet50, and VGG, and multiple image training sizes (321 $\times$ 321, 473 $\times$ 473). We used dilated convolution as the backbone to encode the features, maintaining the feature resolution, as done in previous works~\cite{zhang2019canet, crnet, pfenet,liu2024gaussian,liu2025physically}. During the training process, we used data augmentation techniques such as random crop, random scale, and random flip.

We conducted ablation experiments with a Resnet-50 backbone and $473 \times 437$ training image size, unless specified otherwise.

\subsection{Datasets and Evaluation Metric}
\textbf{PASCAL VOC 2012.} 
The PASCAL-$5^i$ dataset~\cite{everingham2010pascal} consists of 20 categories, with 10582 images for training, 1449 for validation, and 1456 for testing. To ensure comparability with previous work, we follow the data split used in \cite{pfenet} and divide the dataset into training and testing sets. Specifically, we use a cross-validation approach where we divide the 20 object categories into 4 folds, with three for training and one for testing. During training, we use a batch size of 4 and an initial learning rate of 0.0025. We train our network for 200 epochs.

\input{Figures/visualize_coco}
\input{Figures/visulize_fail_cases}

\textbf{MS COCO.}
One limitation of PASCAL-$5^i$ is that it involves only 20 categories, which may not be enough to evaluate the model's ability to perform few-shot segmentation tasks. To address this, we conduct cross-validation experiments on the larger MS COCO dataset, which contains more categories and images. Specifically, COCO 2014~\cite{coco} includes 80 object categories with 82783 training images and 40504 validation images. Similar to previous works~\cite{pfenet}, we divide the 80 categories into four folds, using three folds for training and one fold for testing.

\textbf{Evaluation Metric.} 
In previous works, there exist two evaluation metrics. Shaban\etal~\cite{shaban2017one} report the results with the standard mean Intersection-Over-Union(mIoU). While~\cite{Dong2018FewShotSS,rakelly2018conditional} ignore the categories and report the results by averaging of foreground IoU and background IoU (FBIoU). Following the previous works~\cite{pfenet,zhang2019canet,shaban2017one}, we choose the standard mIoU as our evaluation metric for the following reasons: 1) The unbalanced image distribution (\eg, in the PASCAL VOC test dataset, class sheep contains 49 images while the class person contains 378 images). 2) The score of the background IoU is very high for small objects, which will fail to evaluate the model's capability. Nevertheless, we still compare the previous state-of-the-art methods with both evaluation metrics. The evaluation metrics are calculated as follows:

\begin{equation}
    IoU = \frac{Intersection}{Union} = \frac{TP}{TP + FP + FN}, 
    \label{equal:iou}
\end{equation}

\begin{equation}
    mIoU = \frac{1}{n} \sum_{1}^{n}(IoU_{n}),
\end{equation}

\begin{equation}
    FBIoU = \frac{1}{2}(IoU_{fg} + IoU_{bg}).
\end{equation}
The TP denotes true positive, FP denotes false positive, FN denotes false negative, $n$ denotes the classes' number. The standard mIoU is calculated by averaging the IoU of all classes. The $IoU_{fg}$ is calculated with equation~\ref{equal:iou}, which ignores the object categories, and $IoU_{bg}$ is calculated in the same way but reverses the foreground and background. FBIoU average the $IoU_{fg}$ and the $IoU_{bg}$.

\subsection{Comparisons with state-of-the-art}
\textbf{PASCAL VOC 2012.} 
To ensure a fair comparison with previous state-of-the-art (SOTA) methods, we conducted experiments using multiple backbone models (VGG16, Resnet50, Resnet101) and training image sizes (321 $\times$ 321 and 473 $\times$ 473). The choice of backbone and image size can significantly impact the final performance of a model.

Table~\ref{sota_voc} displays the mean intersection-over-union (mIoU) results of various methods on the PASCAL-$5^i$ dataset. Our proposed method achieved the best results across different backbone models and image sizes. Notably, our approach outperforms the previous SOTA method, PFENet~\cite{pfenet}, by 9.4\% and 3.0\% mIoU in the 1-shot and 5-shot settings, respectively, when using the VGG-16 backbone. Even when compared to FWBF~\cite{fast-weight}, which utilizes a larger training size, our method still outperforms it.
For Resnet-101, our approach yields a 3.3 mIoU score improvement over PFENet~\cite{pfenet} in the 5-shot setting. These results demonstrate the effectiveness and robustness of our method across different backbones and image sizes, and confirm that our approach achieves superior performance over previous SOTA methods.

\textbf{MS COCO.}
Table~\ref{sota_coco} shows the comparison of our proposed QGNet with previous state-of-the-art methods in terms of the standard mIoU. In the large-scale MS COCO dataset experiments, PFENet used a ResNet-101 backbone with an image size of 641 $\times$ 641 for training. However, due to GPU memory constraints, we trained our model with multiple backbones (ResNet-50, ResNet-101, VGG) and two different training sizes (473 $\times$ 473 and 321 $\times$ 321). Despite this limitation, our proposed QGNet outperformed the previous state-of-the-art (PFENet) by 5.9\% and 7.4\% in the 1-shot and 5-shot settings, respectively, using the ResNet-50 backbone.

It is worth noting that even though previous methods, such as PFENet~\cite{pfenet} and FWBF~\cite{fast-weight}, used larger training image sizes (e.g., 641 $\times$ 641 or 512 $\times$ 512), our method with the same backbone (e.g., ResNet-101) still outperforms all those methods with smaller training sizes. In particular, our method with a training size of 473 $\times$ 473 achieves a 10.1\% mIoU improvement in Fold-3 with a 1-shot setting. For the 1-shot mean performance, our method outperforms PFENet by 8.1\% mIoU and FWBF by more than 19.3\% mIoU score. For the 5-shot setting, our method outperforms the previous state-of-the-art with a mean 7.8\% mIoU with a smaller training size. These results demonstrate the superiority of our proposed QGNet.

In addition, we visualize some testing results on fold 0 of the PASCAL-$5^i$ dataset in Figure~\ref{fig:visulization} and some testing results on the COCO dataset in Figure~\ref{fig:visual-coco}.

\subsection{Ablation studies}
Our method introduces a novel self-correspondence module that generates a prior region map for query images independently. To demonstrate the effectiveness of our proposed global and local contrastive losses for prior extractor training, we conducted ablation experiments. We also conducted experiments to identify the best approach for local patch generation during local contrastive loss training. To gain further insights into the impact of the generated prior region maps on query mask prediction, we analyzed the overlap between the prior region maps and the guided region maps.

\textbf{Prior Extractor Training.}
We performed ablation experiments to analyze which patch generation method produces better performance for prior extractor training. Our results, presented in Table~\ref{table:ablation-prio-generation}, demonstrate that we obtain improved query mask predictions when using Slic~\cite{slic} as the patch generation method. Figure~\ref{fig:visulize_slic} illustrates that the patches generated by Slic are larger than those generated by Felzenszwalb’s method and each patch is likely capturing a distinct object pattern. We posit that an appropriate patch size may enhance the prior extractor's capacity to learn a more distinctive predetermined prior. 

Furthermore, in Table~\ref{table:ablation-prio-generation}, we demonstrate our ablation experiments. These results indicate that training the prior extractor for 1000 epochs with the Slic method achieves the best performance.

\textbf{Effectiveness of global and local contrastive losses.}
We conducted ablation experiments to evaluate the effectiveness of our proposed global and local contrastive losses. The results, presented in Table~\ref{table:ablation-global-local}, indicate that the combination of both losses yields the best performance. This suggests that using global and local contrastive losses during prior extractor training can help store more discriminative prior for query mask prediction.

\textbf{Recall for region maps.}
Furthermore, we conducted additional experiments to assess the recall ability of the guided region map and prior region map with the ground truth query segmentation mask. We defined the object regions $R$ with a threshold $\alpha$, such that if the Intersection over Union (IoU) between the ground truth query mask and the object region is greater than $\alpha$, the region is considered a true positive. 

It is worth noting that the `Prior' map contains more information compared to the `Guided Region Map.' During the generation of the guided region map, we filtered out background information and retained only the target information. This filtering process may lead to the loss of some important background information, even though the background can provide valuable information for few-shot segmentation.

\begin{equation}
R = \begin{cases}
    1, & \text{if $V(x,y) > \alpha$},\\
    0, & \text{otherwise}.
  \end{cases} 
\label{equation:masks}
\end{equation}
$V$ denotes the normalized values of the feature map. 

The recall of the region maps with groundtruth mask is calculated as $ Recall = (R \cap R_{GT}) /R_{GT}$. Here $R_{GT}$ indicates the groundtruth mask region. $Recall$ denotes the overlap between the region map and groundtruth mask divided by the groundtruth mask. The higher the $Recall$, the more object regions of the region map can cover the groundtruth mask. In Table~\ref{ablation_iou}, the recall of the prior region map (denoted as $Recall_{p, gt}$) is larger than the recall of the guided region map (denoted as $Recall_{g, gt}$), which suggests that the prior region map could cover more potential object regions.

\subsection{Failure case analysis}
In this section, we aim to analyze the challenging cases that our model fails to correctly identify on the COCO dataset. Figure~\ref{fig:fail-cases} shows some of these instances where our model is unable to distinguish between people and Teddy bears, horses, and people due to their similar patterns or colors. These scenarios are difficult to distinguish without the use of semantic information. Furthermore, our model also struggles to locate small objects in the images. When the query information is too complex, our model can only identify a small part of the human being, indicating that more advanced techniques may be required to improve performance in such challenging cases.

%% file: Tables/sota_voc.tex
\begin{table*}
\centering
\caption{1-shot and 5-shot mIoU results on PASCAL-5$^i$ dataset. The training size and backbone used by each method are listed. 
Our QGNet outperforms the state-of-the-art under all the experiment settings. The results reported with mIoU(\%)}
\label{sota_voc}
\resizebox{1\linewidth}{!}{
\begin{tabular}{c|c|c|cccc|c|cccc|c}
\hline
\toprule[1.5pt]
{}                          & {}                                & {}                           & \multicolumn{5}{c|}{{1-shot}}                                                                                                       & \multicolumn{5}{c}{{5 shot}}                                                                                                       \\ \cline{4-13} 
\multirow{-2}{*}{{Methods}} & \multirow{-2}{*}{{Training Size}} & \multirow{-2}{*}{{Backbone}} & {Fold-0} & {Fold-1} & {Fold-2} & {Fold-3} & {Mean}          & {Fold-0} & {Fold-1} & {Fold-2} & {Fold-3} & {Mean}          \\ 
\midrule[0.5pt]

{OSLSM}                      & {-}                       & {VGG 16}                        & 33.6 & 55.3 & 40.9 & 33.5 & 40.8  & 35.9 & 58.1 & 42.7 & 39.1 & 43.9          \\ 
{co-FCN}                      & {-}                       & {VGG 16}                         & 36.7 & 50.6 & 44.9 & 32.4 & 41.1  & 37.5 & 50.0 & 44.1 & 33.9 & 41.4         \\

{AMP-2}                      & {-}                       & {VGG 16}                        & 41.9 & 50.2 & 46.7 & 34.7 & 43.4  & 40.3 & 55.3 & 49.9 & 40.1 & 46.4 \\    

{PFENet}                      & {473  $\times$ 473}                       & {VGG 16}          & 42.3 & 58.0 & 51.1 & 41.2 & 48.1 & 51.8 & 64.6 & 59.8 & 46.5 & 55.7         \\

{PANet}                     & {417 $\times$ 417 }                       & {VGG 16}                  & {42.3}   & {58.0}   & {51.1}   & {41.2}   & {48.1}          & {51.8}   & {64.6}   & {59.8}   & {46.5}   & {55.7}          \\ 

{FWBF}                     & {512 $\times$ 512}                       & {VGG 16}                  & {47.0}   & {59.6}   & {52.6}   & {48.3}   & {51.9}          & {50.9}   & {62.9}   & {56.5}   & {50.1}   & {55.1}          \\ 

SS-PFENet & - & {VGG 16} & 54.5 & 67.4 &63.4 &54.0 &59.8 &56.9 & 70.0 & 68.3 & 62.1 & 64.3 \\ 

\midrule[0.5pt]

{Ours}                      & {321  $\times$  321}                       & {VGG 16}                        &52.9 &65.0 &50.7 &51.6 & 55.0        &56.8 &67.3 &51.2 &58.2 & 58.3         \\ 
{Ours}                      & {473 $\times$ 473}                       & {VGG 16}                        &58.6 &67.2 &52.3 &52.0 & 57.5         &58.7 &68.4 &52.6 &55.0 & 58.7          \\
\midrule[0.5pt] \midrule[0.5pt]

{CANet}                     & {321 $\times$ 321}                       & {ResNet 50}                  & {52.5}   & {65.9}   & {51.3}   & {51.9}   & {55.4}          & {55.5}   & {67.8}   & {51.9}   & {53.2}   & {57.1}          \\ 
{PGNet}                     & {321 $\times$ 321}                       & {ResNet 50}                  & {56.0}   & {66.9}   & {50.6}   & {50.4}   & {56.0}          & {54.9}   & {67.4}   & {51.8}   & {53.0}   & {56.8}          \\ 
{CRNet}                     & {321 $\times$ 321}                       & {ResNet 50}                  & {-}      & {-}      & {-}      & {-}      & {55.7}          & {-}      & {-}      & {-}      & {-}      & {58.8}          \\ 
{PMMs}                       & {321 $\times$ 321}                       & {ResNet 50}                  & {55.2}   & {66.9}   & {52.6}   & {50.7}   & {56.3}          & {56.3}   & {67.3}   & {54.5}   & {51.0}   & {57.3}          \\ 
{PPNet}                     & {417 $\times$ 417}                       & {ResNet 50}                  & {47.8}   & {58.8}   & {53.8}   & {45.6}   & {51.5}          & {58.4}   & {67.8}   & {64.9}   & {56.7}   & {62.0}          \\ 

{PANet}             & {417 $\times$ 417 }                       & {ResNet 50}                  & 44.0 & 57.5 & 50.8 & 44.0 & 49.1  & 55.3 & 67.2 & 61.3 & 53.2 & 59.3    \\ 

{PFENet}                    & {473 X 473}                       & {ResNet 50}               & {61.7}   & {69.5}   & {55.4}   & {56.3}   & {60.8}          & {63.1}   & {70.7}   & {55.8}   & {57.9}   & {61.9}          \\

SCL (CANet)  & - & {ResNet 50} & 56.8 & 67.3 &53.5 &52.5 &57.5 &59.5 &68.5 &54.9 &53.7 &59.2 \\

SCL (PFENet) & - & {ResNet 50} &63.0 &70.0 &56.5 &57.7 &61.8 &64.5 &70.9 &57.3 &58.7 &62.9 \\

SS-PFENet& - & {ResNet 50} & 58.9 & \textbf{69.9} & \textbf{66.4} &57.7 &\textbf{63.2} & 61.4 & \textbf{75.0} & \textbf{70.5} & \textbf{67.7} & \textbf{68.6} \\ 

\midrule[0.5pt]

{Ours}                      & {321 $\times$ 321}                       & {ResNet 50}                        &57.9 &67.2 &52.4 &55.5 & 58.2        &59.2 &69.4 &53.0 &64.5 & 61.5         \\ 

{Ours}                      & {473 $\times$ 473}                       & {ResNet 50}               & \textbf{63.4 }   & {69.4}   & {55.1}   & \textbf{58.4}   & {61.6} & {64.7}  & {71.0}  & {53.6}  & {61.6}  & {62.8} \\

\midrule[0.5pt] \midrule[0.5pt]

{FWBF}                     & {512 $\times$ 512}               & {ResNet 101}      & 51.3 & 64.5 & 56.7 & 52.2 & 56.2 & 54.8 & 67.4 & 62.2 & 55.3 & 59.9         \\ 
{PPNet}                      & {417 $\times$ 417}                       & {ResNet 101}      & 52.7 & 62.8 & 57.4 & 47.7 & 55.2 & 60.3 & 70.0 & {69.4} & {60.7} & 65.1     \\
{DAN}                      & {-}                       & {ResNet 101}                  & 54.7 & 68.6 & 57.8 & 51.6 & 58.2 & 57.9 & 69.0 & 60.1 & 54.9 & 60.5         \\ 
{PFENet}                    & {473 X 473}                    & {ResNet 101}               & 60.5 & 69.4 & 54.4 & 55.9 & 60.1  & 62.8 & 70.4 & 54.9 & 57.6 & 61.4          \\

\midrule[0.5pt]
{Ours}                      & {321 $\times$ 321}                       & {ResNet 101}                       &57.6 &67.5 &53.0 &53.6 &57.9          &58.6 &66.5 &53.3 &53.6 & 58.0        \\ 
{Ours}                      & {473 X 473}                       & {ResNet 101}                       &60.3 &69.3 &53.3 &57.4 & 60.1         &\textbf{67.6} &71.8 &55.1 &{64.4} & {64.7}         \\

\bottomrule[1.5pt]
\end{tabular}
}
\vspace{+0.36cm}
\end{table*}

%% file: Figures/visualize_slic.tex
\begin{figure*}[t]
\centering
    \includegraphics[width=0.75\linewidth]{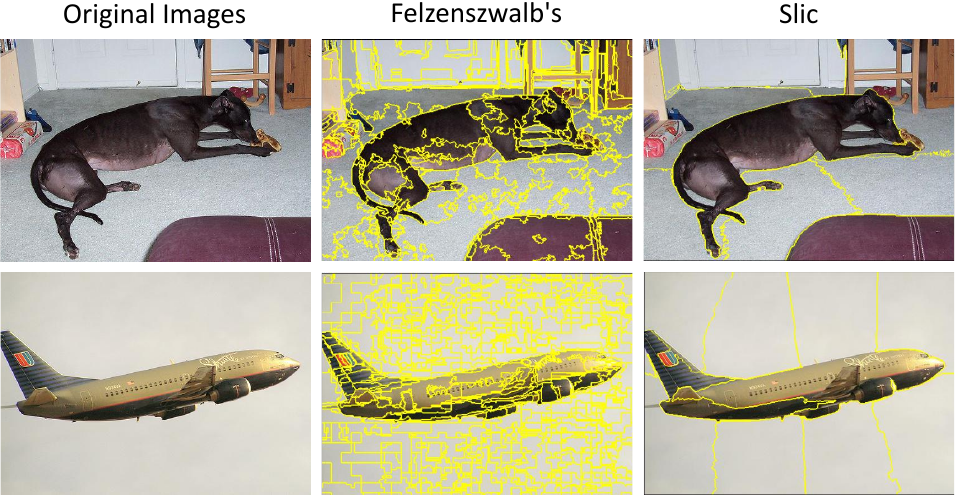}
        \vspace{-0.26cm}
    \caption{Visualization of the local patches generated by Felzenszwalb's method and Slic.}
    % \vspace{+0.16cm}
    \label{fig:visulize_slic}
\end{figure*}

%% file: Figures/visualize.tex
\begin{figure*}[t]
\centering
    \includegraphics[width=1\linewidth]{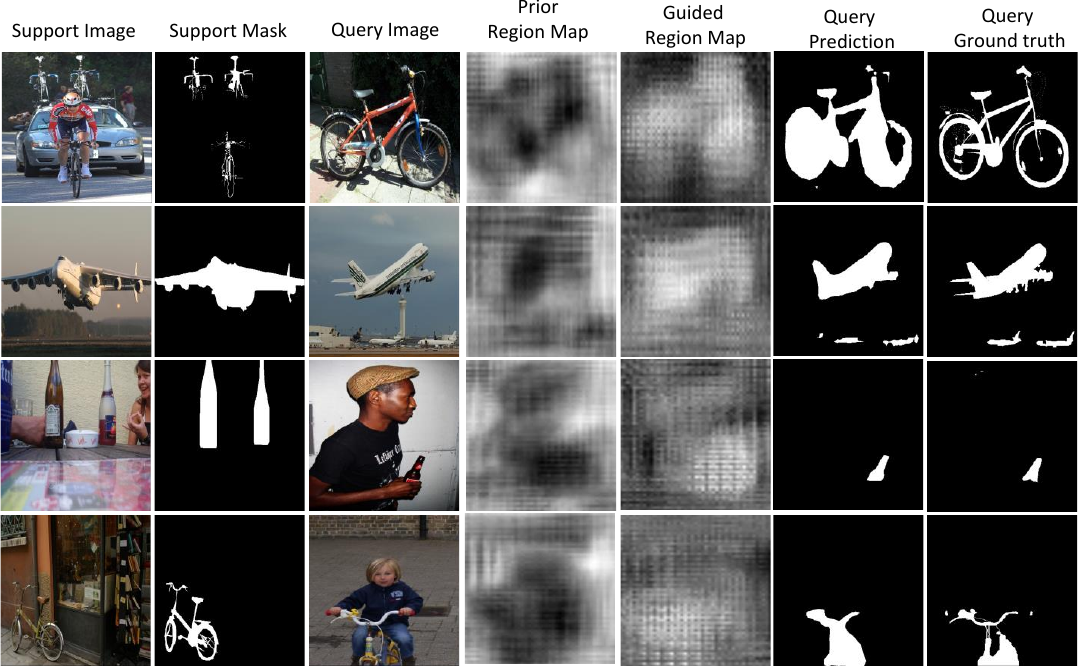}
    \vspace{-0.26cm}
    \caption{Visualization results for guided region map, prior region map, and the query prediction generated by our proposed QGNet on PASCAL-$5^i$ dataset.}
    \label{fig:visulization}
\end{figure*}

%% file: Tables/fbiou-voc.tex
\begin{table}[t]
\centering
\small
\caption{Comparison with the state-of-the-art methods under the 1-shot and 5-shot setting. 
The results reported on PASCAL VOC 2012 dataset with FBIoU(\%).}
\resizebox{0.75\columnwidth}{!}
{

\begin{tabular}{lcc}
\toprule[1pt]
Methods & 1-shot (\%) & 5-shot (\%) \\
\hline
OSLM      & 61.3    & 61.5          \\
co-fcn    & 60.9      & 60.2          \\
sg-one     & 63.1  & 65.9          \\
R-DFCN    & 60.9   & 66.0          \\
PL            & 61.2     & 62.3          \\

A-MCG        & 61.2    & 62.2          \\
CANet        & 66.2    & 69.6          \\
CRNet     & 66.8    & 71.5     \\
% PFENet \cite{pfenet} & 73.3    & 73.9         \\
\hline
Ours     & \textbf{72.2}    & \textbf{73.9}     \\
\bottomrule[1pt]
\end{tabular}

}
\vspace{+0.1cm}

\label{table:voc-fbiou}
\end{table}

%% file: Tables/sota_coco.tex
\begin{table*}
\centering
\caption{1-shot and 5-shot mIoU results on COCO dataset. The results of CANet* is obtained from~\cite{pmm}.} 
\centering
\label{sota_coco}
\resizebox{1\linewidth}{!}{
\begin{tabular}{c|c|c|cccc|c|cccc|c}
\toprule[1.5pt]
{}                          & {}                                & {}                           & \multicolumn{5}{c|}{{1-shot}}                                                                                                       & \multicolumn{5}{c}{{5 shot}}                                                                                                       \\ \cline{4-13} 
\multirow{-2}{*}{{Methods}} & \multirow{-2}{*}{{Training Size}} & \multirow{-2}{*}{{Backbone}} & {Fold-0} & {Fold-1} & {Fold-2} & {Fold-3} & {Mean}          & {Fold-0} & {Fold-1} & {Fold-2} & {Fold-3} & {Mean}          \\  \midrule[0.5pt]
PANet   &     417 $\times$ 417          & VGG 16     & -     & -                              & -    & -     & 20.9    & -    & -    & -    & -     & 29.7 \\ 
FWBF      & {512 $\times$ 512}      & {VGG 16}                        & {18.4}   & {16.7}   & {19.6}   & {25.4}   & {20.0}          & {20.9}   & {19.2}   & {21.9}   & {28.4}   & {22.6}          \\ 
PFENet                      & {473 $\times$ 473}                       & {VGG 16}                       & 33.4 & 36.0 & 34.1 & 32.8 & 34.1  & 35.9 & 40.7 & 38.1 & 36.1 & 37.7   \\

SS-PANet & - & {VGG 16}                       &29.8 &21.2 &26.5 &28.5 &26.2 &36.7 &41.0 &37.6 &35.6 &37.7 \\

SS-PFENet & - & {VGG 16} &35.6 &39.2 &37.6 &37.3 &37.5 & 40.4 & 45.8 & 40.3 & 40.7 & 41.8 \\

\midrule[0.5pt]
{Ous}                      & {321  $\times$ 321}                       & {VGG 16}                      &32.2 &35.2 &32.2 &31.9 &32.8     
&33.8 &38.5 &37.3 &36.4 &36.5        \\
{Ous}                      & {473  $\times$ 473}                       & {VGG 16}                        &32.4 &38.1 &35.7 &32.7 & 34.7         &33.4 &42.2 &39.5 &37.6 &38.2          \\

\midrule[0.5pt] \midrule[0.5pt]

CANet*   & 321 $\times$ 321     & ResNet 50  & 25.1  & 30.3                           & 24.5 & 24.7  & 26.1    & 26.0 & 32.4 & 26.1 & 27.0 & 27.9 \\ 
PMMs   &   321 $\times$ 321  & ResNet 50  & 29.5  & 36.8 & 29.0 & 27.0  & 30.6    & 33.8 & 42.0 & 33.0 & 33.3  & 35.5 \\ 

PPNet    & -    & ResNet 50    & 28.1 & 30.8 & 29.5 & 27.7 & 29.0 & 39.0 & 40.8 & 37.1 & 37.3 & 38.5  \\
                
RPMM    &   321 $\times$ 321 & ResNet 50      & 29.5 & 36.8 & 28.9 & 27.0 & 30.6  & 33.8 & 42.0 & 33.0 & 33.3 & 35.5   \\    
                
PFENet      &  473 $\times$ 473                       &  ResNet 50   & 36.5 & 38.6 & {34.5} & {33.8} & {35.8} & 36.5 & 43.3 & 37.8 & 38.4 & 39.0 \\

\midrule[0.5pt] 

{Ous}                      & {321 X 321}                       & {ResNet 50}                        &34.4 &38.5 &34.6 &33.2 & 35.2        &38.9 &46.2 &39.1 &38.9 & 40.7          \\

Ours    &  473  $\times$ 473       & ResNet 50  & {36.7}  & {41.4}                           & {38.7} & {36.6}  & {38.3}    & {41.5} & \textbf{48.1} & {46.3} & \textbf{43.6}  & {44.8} \\

\midrule[0.5pt] \midrule[0.5pt]

DAN & -    & ResNet 101  & - & - & - & - & 24.4 & - & - & - & - & 29.6  \\
FWBF      &    512 $\times$ 512  & ResNet 101 & 19.9  & 18.0    & 21.0 & 28.9  & 21.2    & 19.1 & 21.5 & 23.9 & 30.1  & 23.7 \\ 
PFENet  & 641 $\times$ 641     & ResNet 101 & 34.3  & 33.0 & 32.3 & 30.1  & 32.4    & 38.5 & 38.6 & 38.2  & 34.3  & 37.4 \\ 

SCL (PFENet) & - & ResNet 101 & 36.4  & 38.6  &37.5  &35.4  &37.0  &38.9  &40.5  &41.5  &38.7  &39.9 \\

\midrule[0.5pt] \midrule[0.5pt]

{Ous}                      & {321  $\times$ 321}                       & {ResNet 101}                       &33.0 &38.4 &36.8 &32.8 & 35.2         &38.9 &46.7 &43.1 &40.4 & 42.3          \\
{Ous}                      & {473  $\times$ 473}                       & {ResNet 101}                       &\textbf{39.0} &\textbf{42.6} &\textbf{40.5} &\textbf{40.2} & \textbf{40.5}        &\textbf{43.8} &46.8 &\textbf{47.5} &42.6 & \textbf{45.2 }      \\

\bottomrule[1.5pt]
\end{tabular}
}
\vspace{+0.46cm}
\end{table*}

%% file: Tables/fbiou-coco.tex
\begin{table}[t]
\centering
\small
\caption{Comparison with the state-of-the-art methods under the 1-shot and 5-shot setting. Our proposed network outperforms all previous methods and achieves new state-of-the-art performance. The results reported on MS COCO dataset with FBIoU(\%).}
\resizebox{0.75\columnwidth}{!}
{

\begin{tabular}{lcc}
\toprule[1pt]
Method & 1-shot (\%) & 5-shot (\%) \\
\hline
PANet       & 59.2    & 63.5          \\
A-MCG  & 52.0      & 54.7         \\
PFENet & 58.6      & 61.9         \\
DAN  & 62.3      & 63.9   \\
\hline
Ours     & \textbf{62.4}    & \textbf{66.2}     \\
\bottomrule[1pt]
\end{tabular}
}

 \vspace{+0.1cm}

\label{table:coco-fbiou}
\end{table}

%% file: Tables/ablation.tex
\begin{table}
\centering
\caption{Ablation studies on PASCAL-5$^i$ dataset about the training epochs and the local patch generation methods for prior extractor training. The results reported with mIoU(\%)}
\label{table:ablation-prio-generation}
\resizebox{\linewidth}{!}{
\begin{tabular}{l|c|cccc|c}
\toprule[1.5pt]
\multirow{2}{*}{Methods} & \multirow{2}{*}{Number of Epochs} & \multicolumn{5}{c}{1 shot}              \\ \cline{3-7} 
                         &                                   & fold 0 & fold 1 & fold 2 & fold 3 & mean \\ \midrule[0.5pt]
Felzenszwalb      & 1000                              & 62.6   & 68.9   & 54.9   & 56.7   & 60.8 \\ 
Felzenszwalb        & 2000                              & 62.7   & 69.3   & 54.7   & 56.8   & 60.9 \\ \hline
Slic                   & 1000                              & {63.4 }   & {69.4}   & {55.1}   & {58.4}   & {61.6} \\ 
Slic                     & 2000                              & 62.4   & 69.4   & 55.2   & 57.0   & 61.0 \\ \bottomrule[1.5pt]
\end{tabular}
}
\vspace{+0.16cm}
\end{table}

%% file: Tables/ablation_global_local.tex
\begin{table}
\centering
\caption{Ablation studies on PASCAL-5$^i$ dataset about global and local contrastive learning. The results reported with mIoU(\%)}
\label{table:ablation-global-local}
\resizebox{1\linewidth}{!}{
\begin{tabular}{c|c|cccc|c}
\toprule[1.5pt]
\multicolumn{2}{l|}{Contrastive learning type:} & \multicolumn{5}{c}{1 shot}              \\ \hline
local         & global        & fold 0 & fold 1 & fold 2 & fold 3 & mean \\ \midrule[0.5pt]
$\surd$             &               &63.2    &69.3  &54.9  &57.4     &  61.2    \\ 
              & $\surd$             & 62.2       &69.7        &54.5        &57.5        &60.9      \\ 
$\surd$             & $\surd$   &   {63.4 }   &{69.4}   &{55.1}   &{58.4}   &{61.6}         \\ \bottomrule[1.5pt]
\end{tabular}
\vspace{+0.36cm}
}
\end{table}

%% file: Tables/ablation_iou.tex
\begin{table*}[]
\centering
\caption{Ablation studies on PASCAL-$5^i$ fold-0 dataset about the recall for guided region map and prior region map with ground truth query mask. The results reported with mIoU(\%)}
\resizebox{0.8\linewidth}{!}{
\begin{tabular}{l|ccccccccc}
\toprule[1.5pt]
threshold & 0.1 & 0.2 & 0.3 & 0.4 & 0.5 & 0.6 & 0.7 & 0.8 & 0.9 \\ \hline

$Recall_{g,gt}$ & 99.9\% & 99.9\% & 99.3\% & 96.0\% & 85.6\% & 65.6\% & 39.6\% & 15.5\% & 2.2\% \\ 
$Recall_{p,gt}$ & 99.9\% & 99.9\% & 99.1\% & 96.5\% & 89.7\% & 76.2\% & 54.7\% & 28.2\% & 6.6\% \\ 
\bottomrule[1.5pt]
\end{tabular}
}
\label{ablation_iou}
\vspace{+0.16cm}
\end{table*}

%% file: Figures/visualize_coco.tex
\begin{figure*}[t]
\centering
    \includegraphics[width=1\linewidth]{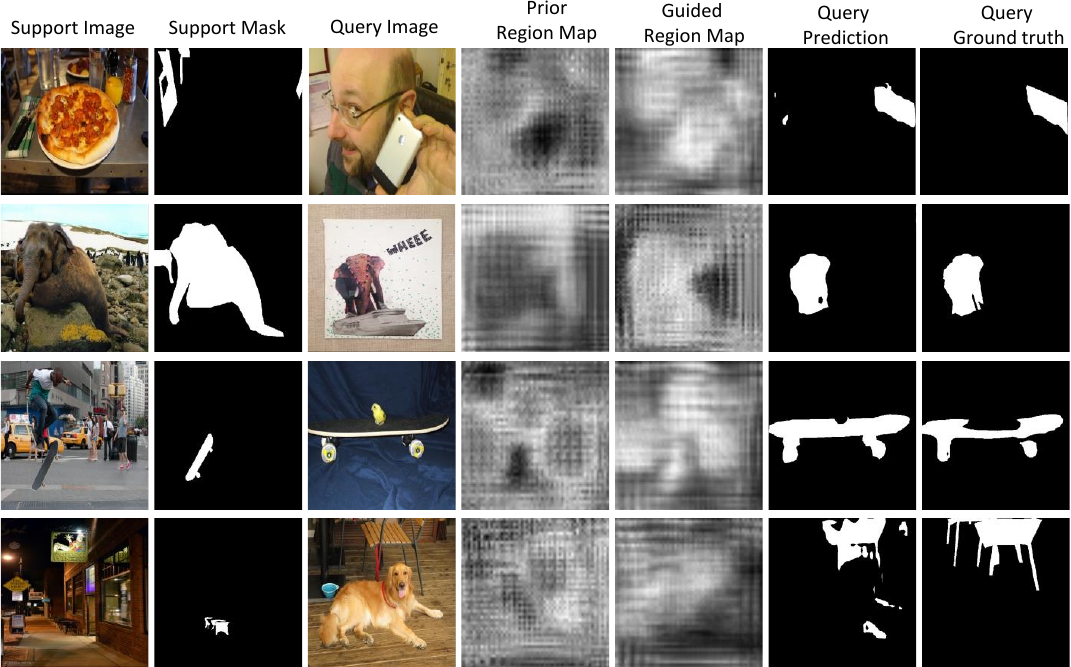}
        \vspace{-0.26cm}
    \caption{Visualization results for guided region map, prior region map, and the query prediction generated by our proposed QGNet on COCO dataset.}
    \label{fig:visual-coco}

\end{figure*}

%% file: Figures/visulize_fail_cases.tex
\begin{figure}[t]
\centering
    \includegraphics[width=1\linewidth]{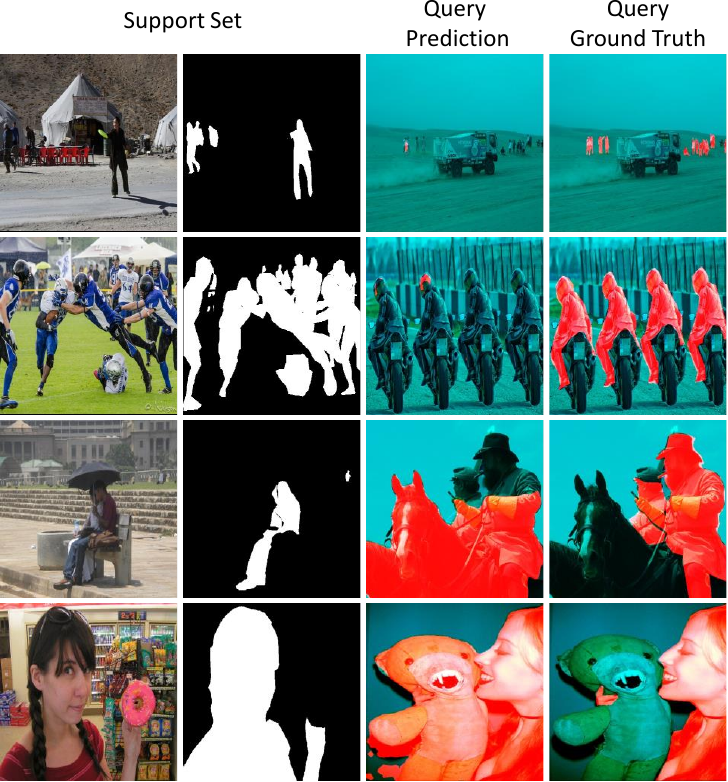}
    \vspace{-0.26cm}
    \caption{The failure cases on COCO dataset.}
    \label{fig:fail-cases}
\end{figure}

%% file: 6_conclusion.tex
\section{conclusion}
This paper proposes the Query Guided Few Shot Segmentation method (QGNet) which employs self-supervised learning to learn query information from unlabeled images. To achieve this, the paper introduces a global-local contrastive loss to train the prior extractor, enabling the query branch to independently extract informative clues from the query image and enhancing cross-interaction between query and support. Experimental results on the PASCAL VOC 2012 and MS COCO datasets demonstrate the effectiveness of the proposed method, achieving new state-of-the-art results.

%% file: cas-dc-template.bbl
\begin{thebibliography}{51}
\expandafter\ifx\csname natexlab\endcsname\relax\def\natexlab#1{#1}\fi
\providecommand{\url}[1]{\texttt{#1}}
\providecommand{\href}[2]{#2}
\providecommand{\path}[1]{#1}
\providecommand{\DOIprefix}{doi:}
\providecommand{\ArXivprefix}{arXiv:}
\providecommand{\URLprefix}{URL: }
\providecommand{\Pubmedprefix}{pmid:}
\providecommand{\doi}[1]{\href{http://dx.doi.org/#1}{\path{#1}}}
\providecommand{\Pubmed}[1]{\href{pmid:#1}{\path{#1}}}
\providecommand{\bibinfo}[2]{#2}
\ifx\xfnm\relax \def\xfnm[#1]{\unskip,\space#1}\fi
%Type = Inproceedings
\bibitem[{Achanta et~al.(2008)Achanta, Estrada, Wils and S{\"u}sstrunk}]{achanta2008salient}
\bibinfo{author}{Achanta, R.}, \bibinfo{author}{Estrada, F.}, \bibinfo{author}{Wils, P.}, \bibinfo{author}{S{\"u}sstrunk, S.}, \bibinfo{year}{2008}.
\newblock \bibinfo{title}{Salient region detection and segmentation}, in: \bibinfo{booktitle}{Computer Vision Systems: 6th International Conference, ICVS 2008 Santorini, Greece, May 12-15, 2008 Proceedings 6}, \bibinfo{organization}{Springer}. pp. \bibinfo{pages}{66--75}.
%Type = Article
\bibitem[{Achanta et~al.(2012)Achanta, Shaji, Smith, Lucchi, Fua and S{\"u}sstrunk}]{slic}
\bibinfo{author}{Achanta, R.}, \bibinfo{author}{Shaji, A.}, \bibinfo{author}{Smith, K.}, \bibinfo{author}{Lucchi, A.}, \bibinfo{author}{Fua, P.}, \bibinfo{author}{S{\"u}sstrunk, S.}, \bibinfo{year}{2012}.
\newblock \bibinfo{title}{Slic superpixels compared to state-of-the-art superpixel methods}.
\newblock \bibinfo{journal}{IEEE transactions on pattern analysis and machine intelligence} \bibinfo{volume}{34}, \bibinfo{pages}{2274--2282}.
%Type = Inproceedings
\bibitem[{Boudiaf et~al.(2021)Boudiaf, Kervadec, Ziko, Piantanida, Ayed and Dolz}]{repri}
\bibinfo{author}{Boudiaf, M.}, \bibinfo{author}{Kervadec, H.}, \bibinfo{author}{Ziko, I.M.}, \bibinfo{author}{Piantanida, P.}, \bibinfo{author}{Ayed, I.B.}, \bibinfo{author}{Dolz, J.}, \bibinfo{year}{2021}.
\newblock \bibinfo{title}{Few-shot segmentation without meta-learning: {A} good transductive inference is all you need?}, in: \bibinfo{booktitle}{Proceedings of the IEEE/CVF Conference on Computer Vision and Pattern Recognition}, pp. \bibinfo{pages}{13979--13988}.
%Type = Article
\bibitem[{Chaitanya et~al.(2020)Chaitanya, Erdil, Karani and Konukoglu}]{chaitanya2020contrastive}
\bibinfo{author}{Chaitanya, K.}, \bibinfo{author}{Erdil, E.}, \bibinfo{author}{Karani, N.}, \bibinfo{author}{Konukoglu, E.}, \bibinfo{year}{2020}.
\newblock \bibinfo{title}{Contrastive learning of global and local features for medical image segmentation with limited annotations}.
\newblock \bibinfo{journal}{Advances in neural information processing systems} \bibinfo{volume}{33}, \bibinfo{pages}{12546--12558}.
%Type = Article
\bibitem[{Chen et~al.(2018)Chen, Papandreou, Kokkinos, Murphy and Yuille}]{chen2018deeplab}
\bibinfo{author}{Chen, L.C.}, \bibinfo{author}{Papandreou, G.}, \bibinfo{author}{Kokkinos, I.}, \bibinfo{author}{Murphy, K.}, \bibinfo{author}{Yuille, A.L.}, \bibinfo{year}{2018}.
\newblock \bibinfo{title}{Deeplab: Semantic image segmentation with deep convolutional nets, atrous convolution, and fully connected crfs}.
\newblock \bibinfo{journal}{IEEE transactions on pattern analysis and machine intelligence} \bibinfo{volume}{40}, \bibinfo{pages}{834--848}.
%Type = Inproceedings
\bibitem[{Chen et~al.(2020a)Chen, Kornblith, Norouzi and Hinton}]{simclr}
\bibinfo{author}{Chen, T.}, \bibinfo{author}{Kornblith, S.}, \bibinfo{author}{Norouzi, M.}, \bibinfo{author}{Hinton, G.}, \bibinfo{year}{2020}a.
\newblock \bibinfo{title}{A simple framework for contrastive learning of visual representations}, in: \bibinfo{booktitle}{International conference on machine learning}, \bibinfo{organization}{PMLR}. pp. \bibinfo{pages}{1597--1607}.
%Type = Article
\bibitem[{Chen et~al.(2020b)Chen, Fan, Girshick and He}]{mocov2}
\bibinfo{author}{Chen, X.}, \bibinfo{author}{Fan, H.}, \bibinfo{author}{Girshick, R.}, \bibinfo{author}{He, K.}, \bibinfo{year}{2020}b.
\newblock \bibinfo{title}{Improved baselines with momentum contrastive learning}.
\newblock \bibinfo{journal}{arXiv preprint arXiv:2003.04297} .
%Type = Inproceedings
\bibitem[{Dong et~al.(2021)Dong, Yang, Xu, Huang and Yu}]{dong2021abpnet}
\bibinfo{author}{Dong, K.}, \bibinfo{author}{Yang, W.}, \bibinfo{author}{Xu, Z.}, \bibinfo{author}{Huang, L.}, \bibinfo{author}{Yu, Z.}, \bibinfo{year}{2021}.
\newblock \bibinfo{title}{Abpnet: Adaptive background modeling for generalized few shot segmentation}, in: \bibinfo{booktitle}{ACM MM}, pp. \bibinfo{pages}{2271--2280}.
%Type = Inproceedings
\bibitem[{Dong and Xing(2018)}]{Dong2018FewShotSS}
\bibinfo{author}{Dong, N.}, \bibinfo{author}{Xing, E.}, \bibinfo{year}{2018}.
\newblock \bibinfo{title}{Few-shot semantic segmentation with prototype learning}, in: \bibinfo{booktitle}{British Machine Vision Conference}.
%Type = Inproceedings
\bibitem[{Dong et~al.(2019)Dong, Zhang, Shao and Zhou}]{mdl}
\bibinfo{author}{Dong, Z.}, \bibinfo{author}{Zhang, R.}, \bibinfo{author}{Shao, X.}, \bibinfo{author}{Zhou, H.}, \bibinfo{year}{2019}.
\newblock \bibinfo{title}{Multi-scale discriminative location-aware network for few-shot semantic segmentation}, in: \bibinfo{booktitle}{2019 IEEE 43rd Annual Computer Software and Applications Conference (COMPSAC)}, \bibinfo{organization}{IEEE}. pp. \bibinfo{pages}{42--47}.
%Type = Article
\bibitem[{Everingham et~al.(2010)Everingham, Van~Gool, Williams, Winn and Zisserman}]{everingham2010pascal}
\bibinfo{author}{Everingham, M.}, \bibinfo{author}{Van~Gool, L.}, \bibinfo{author}{Williams, C.K.}, \bibinfo{author}{Winn, J.}, \bibinfo{author}{Zisserman, A.}, \bibinfo{year}{2010}.
\newblock \bibinfo{title}{The pascal visual object classes (voc) challenge}.
\newblock \bibinfo{journal}{International journal of computer vision} \bibinfo{volume}{88}, \bibinfo{pages}{303--338}.
%Type = Article
\bibitem[{Felzenszwalb and Huttenlocher(2004)}]{felzenszwalb2004efficient}
\bibinfo{author}{Felzenszwalb, P.F.}, \bibinfo{author}{Huttenlocher, D.P.}, \bibinfo{year}{2004}.
\newblock \bibinfo{title}{Efficient graph-based image segmentation}.
\newblock \bibinfo{journal}{International journal of computer vision} \bibinfo{volume}{59}, \bibinfo{pages}{167--181}.
%Type = Article
\bibitem[{Grill et~al.(2020)Grill, Strub, Altch{\'e}, Tallec, Richemond, Buchatskaya, Doersch, Pires, Guo, Azar et~al.}]{Bootstrap}
\bibinfo{author}{Grill, J.B.}, \bibinfo{author}{Strub, F.}, \bibinfo{author}{Altch{\'e}, F.}, \bibinfo{author}{Tallec, C.}, \bibinfo{author}{Richemond, P.H.}, \bibinfo{author}{Buchatskaya, E.}, \bibinfo{author}{Doersch, C.}, \bibinfo{author}{Pires, B.A.}, \bibinfo{author}{Guo, Z.D.}, \bibinfo{author}{Azar, M.G.}, et~al., \bibinfo{year}{2020}.
\newblock \bibinfo{title}{Bootstrap your own latent: A new approach to self-supervised learning}.
\newblock \bibinfo{journal}{arXiv preprint arXiv:2006.07733} .
%Type = Inproceedings
\bibitem[{Hadsell et~al.(2006)Hadsell, Chopra and LeCun}]{contrastivelearn}
\bibinfo{author}{Hadsell, R.}, \bibinfo{author}{Chopra, S.}, \bibinfo{author}{LeCun, Y.}, \bibinfo{year}{2006}.
\newblock \bibinfo{title}{Dimensionality reduction by learning an invariant mapping}, in: \bibinfo{booktitle}{2006 IEEE Computer Society Conference on Computer Vision and Pattern Recognition (CVPR'06)}, \bibinfo{organization}{IEEE}. pp. \bibinfo{pages}{1735--1742}.
%Type = Inproceedings
\bibitem[{He et~al.(2020)He, Fan, Wu, Xie and Girshick}]{moco}
\bibinfo{author}{He, K.}, \bibinfo{author}{Fan, H.}, \bibinfo{author}{Wu, Y.}, \bibinfo{author}{Xie, S.}, \bibinfo{author}{Girshick, R.}, \bibinfo{year}{2020}.
\newblock \bibinfo{title}{Momentum contrast for unsupervised visual representation learning}, in: \bibinfo{booktitle}{Proceedings of the IEEE/CVF Conference on Computer Vision and Pattern Recognition}, pp. \bibinfo{pages}{9729--9738}.
%Type = Inproceedings
\bibitem[{Hossain et~al.(2024)Hossain, Siam, Sigal and Little}]{Hossain_2024_CVPR}
\bibinfo{author}{Hossain, M.R.I.}, \bibinfo{author}{Siam, M.}, \bibinfo{author}{Sigal, L.}, \bibinfo{author}{Little, J.J.}, \bibinfo{year}{2024}.
\newblock \bibinfo{title}{Visual prompting for generalized few-shot segmentation: A multi-scale approach}, in: \bibinfo{booktitle}{Proceedings of the IEEE/CVF Conference on Computer Vision and Pattern Recognition (CVPR)}, pp. \bibinfo{pages}{23470--23480}.
%Type = Inproceedings
\bibitem[{Hu et~al.(2021)Hu, Cui and Wang}]{hu2021region}
\bibinfo{author}{Hu, H.}, \bibinfo{author}{Cui, J.}, \bibinfo{author}{Wang, L.}, \bibinfo{year}{2021}.
\newblock \bibinfo{title}{Region-aware contrastive learning for semantic segmentation}, in: \bibinfo{booktitle}{Proceedings of the IEEE/CVF International Conference on Computer Vision}, pp. \bibinfo{pages}{16291--16301}.
%Type = Inproceedings
\bibitem[{Ji et~al.(2019)Ji, Henriques and Vedaldi}]{iic}
\bibinfo{author}{Ji, X.}, \bibinfo{author}{Henriques, J.F.}, \bibinfo{author}{Vedaldi, A.}, \bibinfo{year}{2019}.
\newblock \bibinfo{title}{Invariant information clustering for unsupervised image classification and segmentation}, in: \bibinfo{booktitle}{Proceedings of the IEEE/CVF international conference on computer vision}, pp. \bibinfo{pages}{9865--9874}.
%Type = Inproceedings
\bibitem[{Li et~al.(2021a)Li, Jampani, Sevilla-Lara, Sun, Kim and Kim}]{asgnet}
\bibinfo{author}{Li, G.}, \bibinfo{author}{Jampani, V.}, \bibinfo{author}{Sevilla-Lara, L.}, \bibinfo{author}{Sun, D.}, \bibinfo{author}{Kim, J.}, \bibinfo{author}{Kim, J.}, \bibinfo{year}{2021}a.
\newblock \bibinfo{title}{Adaptive prototype learning and allocation for few-shot segmentation}, in: \bibinfo{booktitle}{Proceedings of the IEEE/CVF Conference on Computer Vision and Pattern Recognition}, pp. \bibinfo{pages}{8334--8343}.
%Type = Inproceedings
\bibitem[{Li et~al.(2020)Li, Wei, Chen, Tai and Tang}]{fss1000}
\bibinfo{author}{Li, X.}, \bibinfo{author}{Wei, T.}, \bibinfo{author}{Chen, Y.P.}, \bibinfo{author}{Tai, Y.W.}, \bibinfo{author}{Tang, C.K.}, \bibinfo{year}{2020}.
\newblock \bibinfo{title}{Fss-1000: A 1000-class dataset for few-shot segmentation}, in: \bibinfo{booktitle}{Proceedings of the IEEE/CVF Conference on Computer Vision and Pattern Recognition}, pp. \bibinfo{pages}{2869--2878}.
%Type = Inproceedings
\bibitem[{Li et~al.(2021b)Li, Data, Fu, Hu and Prisacariu}]{bmvc_few_shot}
\bibinfo{author}{Li, Y.}, \bibinfo{author}{Data, G.W.P.}, \bibinfo{author}{Fu, Y.}, \bibinfo{author}{Hu, Y.}, \bibinfo{author}{Prisacariu, V.A.}, \bibinfo{year}{2021}b.
\newblock \bibinfo{title}{Few-shot semantic segmentation with self-supervision from pseudo-classes}, in: \bibinfo{booktitle}{British Machine Vision Conference}.
%Type = Inproceedings
\bibitem[{Lin et~al.(2014)Lin, Maire, Belongie, Hays, Perona, Ramanan, Doll{\'a}r and Zitnick}]{coco}
\bibinfo{author}{Lin, T.Y.}, \bibinfo{author}{Maire, M.}, \bibinfo{author}{Belongie, S.}, \bibinfo{author}{Hays, J.}, \bibinfo{author}{Perona, P.}, \bibinfo{author}{Ramanan, D.}, \bibinfo{author}{Doll{\'a}r, P.}, \bibinfo{author}{Zitnick, C.L.}, \bibinfo{year}{2014}.
\newblock \bibinfo{title}{Microsoft coco: Common objects in context}, in: \bibinfo{booktitle}{ECCV}, pp. \bibinfo{pages}{740--755}.
%Type = Article
\bibitem[{Liu et~al.(2025)Liu, Lou, Wang, Zhou, Cheng and Yang}]{liu2025physically}
\bibinfo{author}{Liu, W.}, \bibinfo{author}{Lou, J.}, \bibinfo{author}{Wang, X.}, \bibinfo{author}{Zhou, W.}, \bibinfo{author}{Cheng, J.}, \bibinfo{author}{Yang, X.}, \bibinfo{year}{2025}.
\newblock \bibinfo{title}{Physically-guided open vocabulary segmentation with weighted patched alignment loss}.
\newblock \bibinfo{journal}{Neurocomputing} \bibinfo{volume}{614}, \bibinfo{pages}{128788}.
%Type = Article
\bibitem[{Liu et~al.(2024a)Liu, Wang, Wang, Cheng, Liu and Yang}]{liu2024gaussian}
\bibinfo{author}{Liu, W.}, \bibinfo{author}{Wang, X.}, \bibinfo{author}{Wang, L.}, \bibinfo{author}{Cheng, J.}, \bibinfo{author}{Liu, F.}, \bibinfo{author}{Yang, X.}, \bibinfo{year}{2024}a.
\newblock \bibinfo{title}{Gaussian mixture based evidential learning for stereo matching}.
\newblock \bibinfo{journal}{arXiv preprint arXiv:2408.02796} .
%Type = Article
\bibitem[{Liu et~al.(2024b)Liu, Wu, Zhao, Fang, Foo, Cheng and Lin}]{liu2024harmonizing}
\bibinfo{author}{Liu, W.}, \bibinfo{author}{Wu, Z.}, \bibinfo{author}{Zhao, Y.}, \bibinfo{author}{Fang, Y.}, \bibinfo{author}{Foo, C.S.}, \bibinfo{author}{Cheng, J.}, \bibinfo{author}{Lin, G.}, \bibinfo{year}{2024}b.
\newblock \bibinfo{title}{Harmonizing base and novel classes: A class-contrastive approach for generalized few-shot segmentation}.
\newblock \bibinfo{journal}{International Journal of Computer Vision} \bibinfo{volume}{132}, \bibinfo{pages}{1277--1291}.
%Type = Article
\bibitem[{Liu et~al.(2022a)Liu, Zhang, Ding, Hung and Lin}]{liu2022few}
\bibinfo{author}{Liu, W.}, \bibinfo{author}{Zhang, C.}, \bibinfo{author}{Ding, H.}, \bibinfo{author}{Hung, T.Y.}, \bibinfo{author}{Lin, G.}, \bibinfo{year}{2022}a.
\newblock \bibinfo{title}{Few-shot segmentation with optimal transport matching and message flow}.
\newblock \bibinfo{journal}{IEEE Transactions on Multimedia} \bibinfo{volume}{25}, \bibinfo{pages}{5130--5141}.
%Type = Inproceedings
\bibitem[{Liu et~al.(2020a)Liu, Zhang, Lin and Liu}]{crnet}
\bibinfo{author}{Liu, W.}, \bibinfo{author}{Zhang, C.}, \bibinfo{author}{Lin, G.}, \bibinfo{author}{Liu, F.}, \bibinfo{year}{2020}a.
\newblock \bibinfo{title}{Crnet: Cross-reference networks for few-shot segmentation}, in: \bibinfo{booktitle}{Proceedings of the IEEE/CVF Conference on Computer Vision and Pattern Recognition}, pp. \bibinfo{pages}{4165--4173}.
%Type = Article
\bibitem[{Liu et~al.(2022b)Liu, Zhang, Lin and Liu}]{liu2022crcnet}
\bibinfo{author}{Liu, W.}, \bibinfo{author}{Zhang, C.}, \bibinfo{author}{Lin, G.}, \bibinfo{author}{Liu, F.}, \bibinfo{year}{2022}b.
\newblock \bibinfo{title}{Crcnet: Few-shot segmentation with cross-reference and region--global conditional networks}.
\newblock \bibinfo{journal}{International Journal of Computer Vision} \bibinfo{volume}{130}, \bibinfo{pages}{3140--3157}.
%Type = Inproceedings
\bibitem[{Liu et~al.(2020b)Liu, Zhang, Zhang and He}]{ppnet}
\bibinfo{author}{Liu, Y.}, \bibinfo{author}{Zhang, X.}, \bibinfo{author}{Zhang, S.}, \bibinfo{author}{He, X.}, \bibinfo{year}{2020}b.
\newblock \bibinfo{title}{Part-aware prototype network for few-shot semantic segmentation}, in: \bibinfo{booktitle}{European Conference on Computer Vision}, \bibinfo{organization}{Springer}. pp. \bibinfo{pages}{142--158}.
%Type = Inproceedings
\bibitem[{Long et~al.(2015)Long, Shelhamer and Darrell}]{long2015fully}
\bibinfo{author}{Long, J.}, \bibinfo{author}{Shelhamer, E.}, \bibinfo{author}{Darrell, T.}, \bibinfo{year}{2015}.
\newblock \bibinfo{title}{Fully convolutional networks for semantic segmentation}, in: \bibinfo{booktitle}{Proceedings of the IEEE conference on computer vision and pattern recognition}, pp. \bibinfo{pages}{3431--3440}.
%Type = Inproceedings
\bibitem[{Lu et~al.(2021)Lu, He, Zhu, Zhang, Song and Xiang}]{cwt}
\bibinfo{author}{Lu, Z.}, \bibinfo{author}{He, S.}, \bibinfo{author}{Zhu, X.}, \bibinfo{author}{Zhang, L.}, \bibinfo{author}{Song, Y.Z.}, \bibinfo{author}{Xiang, T.}, \bibinfo{year}{2021}.
\newblock \bibinfo{title}{Simpler is better: Few-shot semantic segmentation with classifier weight transformer}, in: \bibinfo{booktitle}{Proceedings of the IEEE/CVF international conference on computer vision}, pp. \bibinfo{pages}{8721--8730}.
%Type = Inproceedings
\bibitem[{Moon et~al.(2023)Moon, Sohn, Zhou, Yoon, Pavlovic, Khan and Kapadia}]{Moon_2023_ICCV}
\bibinfo{author}{Moon, S.}, \bibinfo{author}{Sohn, S.S.}, \bibinfo{author}{Zhou, H.}, \bibinfo{author}{Yoon, S.}, \bibinfo{author}{Pavlovic, V.}, \bibinfo{author}{Khan, M.H.}, \bibinfo{author}{Kapadia, M.}, \bibinfo{year}{2023}.
\newblock \bibinfo{title}{{MSI}: Maximize support-set information for few-shot segmentation}, in: \bibinfo{booktitle}{Proceedings of the IEEE/CVF International Conference on Computer Vision (ICCV)}, pp. \bibinfo{pages}{19266--19276}.
%Type = Inproceedings
\bibitem[{Nguyen and Todorovic(2019)}]{fast-weight}
\bibinfo{author}{Nguyen, K.}, \bibinfo{author}{Todorovic, S.}, \bibinfo{year}{2019}.
\newblock \bibinfo{title}{Feature weighting and boosting for few-shot segmentation}, in: \bibinfo{booktitle}{Proceedings of the IEEE International Conference on Computer Vision}, pp. \bibinfo{pages}{622--631}.
%Type = Article
\bibitem[{Oord et~al.(2018)Oord, Li and Vinyals}]{infoace}
\bibinfo{author}{Oord, A.v.d.}, \bibinfo{author}{Li, Y.}, \bibinfo{author}{Vinyals, O.}, \bibinfo{year}{2018}.
\newblock \bibinfo{title}{Representation learning with contrastive predictive coding}.
\newblock \bibinfo{journal}{arXiv preprint arXiv:1807.03748} .
%Type = Inproceedings
\bibitem[{Peng et~al.(2023)Peng, Tian, Wu, Wang, Liu, Su and Jia}]{Peng_2023_CVPR}
\bibinfo{author}{Peng, B.}, \bibinfo{author}{Tian, Z.}, \bibinfo{author}{Wu, X.}, \bibinfo{author}{Wang, C.}, \bibinfo{author}{Liu, S.}, \bibinfo{author}{Su, J.}, \bibinfo{author}{Jia, J.}, \bibinfo{year}{2023}.
\newblock \bibinfo{title}{Hierarchical dense correlation distillation for few-shot segmentation}, in: \bibinfo{booktitle}{Proceedings of the IEEE/CVF Conference on Computer Vision and Pattern Recognition (CVPR)}, pp. \bibinfo{pages}{23641--23651}.
%Type = Inproceedings
\bibitem[{Rakelly et~al.(2018)Rakelly, Shelhamer, Darrell, Efros and Levine}]{rakelly2018conditional}
\bibinfo{author}{Rakelly, K.}, \bibinfo{author}{Shelhamer, E.}, \bibinfo{author}{Darrell, T.}, \bibinfo{author}{Efros, A.}, \bibinfo{author}{Levine, S.}, \bibinfo{year}{2018}.
\newblock \bibinfo{title}{Conditional networks for few-shot semantic segmentation}, in: \bibinfo{booktitle}{ICLR Workshop}.
%Type = Article
\bibitem[{Shaban et~al.(2017)Shaban, Bansal, Liu, Essa and Boots}]{shaban2017one}
\bibinfo{author}{Shaban, A.}, \bibinfo{author}{Bansal, S.}, \bibinfo{author}{Liu, Z.}, \bibinfo{author}{Essa, I.}, \bibinfo{author}{Boots, B.}, \bibinfo{year}{2017}.
\newblock \bibinfo{title}{One-shot learning for semantic segmentation}.
\newblock \bibinfo{journal}{arXiv preprint arXiv:1709.03410} .
%Type = Article
\bibitem[{Siam and Oreshkin(2019)}]{siam2019adaptive}
\bibinfo{author}{Siam, M.}, \bibinfo{author}{Oreshkin, B.}, \bibinfo{year}{2019}.
\newblock \bibinfo{title}{Adaptive masked weight imprinting for few-shot segmentation}.
\newblock \bibinfo{journal}{arXiv preprint arXiv:1902.11123} .
%Type = Inproceedings
\bibitem[{Siam et~al.(2019)Siam, Oreshkin and Jagersand}]{amp}
\bibinfo{author}{Siam, M.}, \bibinfo{author}{Oreshkin, B.N.}, \bibinfo{author}{Jagersand, M.}, \bibinfo{year}{2019}.
\newblock \bibinfo{title}{Amp: Adaptive masked proxies for few-shot segmentation}, in: \bibinfo{booktitle}{Proceedings of the IEEE International Conference on Computer Vision}, pp. \bibinfo{pages}{5249--5258}.
%Type = Article
\bibitem[{Tian et~al.(2020)Tian, Zhao, Shu, Yang, Li and Jia}]{pfenet}
\bibinfo{author}{Tian, Z.}, \bibinfo{author}{Zhao, H.}, \bibinfo{author}{Shu, M.}, \bibinfo{author}{Yang, Z.}, \bibinfo{author}{Li, R.}, \bibinfo{author}{Jia, J.}, \bibinfo{year}{2020}.
\newblock \bibinfo{title}{Prior guided feature enrichment network for few-shot segmentation}.
\newblock \bibinfo{journal}{Transactions on Pattern Analysis and Machine Intelligence} .
%Type = Inproceedings
\bibitem[{Wang et~al.(2019)Wang, Liew, Zou, Zhou and Feng}]{wang2019panet}
\bibinfo{author}{Wang, K.}, \bibinfo{author}{Liew, J.H.}, \bibinfo{author}{Zou, Y.}, \bibinfo{author}{Zhou, D.}, \bibinfo{author}{Feng, J.}, \bibinfo{year}{2019}.
\newblock \bibinfo{title}{Panet: Few-shot image semantic segmentation with prototype alignment}, in: \bibinfo{booktitle}{Proceedings of the IEEE International Conference on Computer Vision}, pp. \bibinfo{pages}{9197--9206}.
%Type = Inproceedings
\bibitem[{Wu et~al.(2018)Wu, Xiong, Yu and Lin}]{tau}
\bibinfo{author}{Wu, Z.}, \bibinfo{author}{Xiong, Y.}, \bibinfo{author}{Yu, S.X.}, \bibinfo{author}{Lin, D.}, \bibinfo{year}{2018}.
\newblock \bibinfo{title}{Unsupervised feature learning via non-parametric instance discrimination}, in: \bibinfo{booktitle}{Proceedings of the IEEE Conference on Computer Vision and Pattern Recognition}, pp. \bibinfo{pages}{3733--3742}.
%Type = Inproceedings
\bibitem[{Xie et~al.(2021)Xie, Ding, Wang, Zhan, Xu, Sun, Li and Luo}]{xie2021detco}
\bibinfo{author}{Xie, E.}, \bibinfo{author}{Ding, J.}, \bibinfo{author}{Wang, W.}, \bibinfo{author}{Zhan, X.}, \bibinfo{author}{Xu, H.}, \bibinfo{author}{Sun, P.}, \bibinfo{author}{Li, Z.}, \bibinfo{author}{Luo, P.}, \bibinfo{year}{2021}.
\newblock \bibinfo{title}{Detco: Unsupervised contrastive learning for object detection}, in: \bibinfo{booktitle}{Proceedings of the IEEE/CVF International Conference on Computer Vision}, pp. \bibinfo{pages}{8392--8401}.
%Type = Inproceedings
\bibitem[{Xu et~al.(2024)Xu, Lin, Loy, Long, Li and Zhao}]{Xu_2024_ECCV}
\bibinfo{author}{Xu, Q.}, \bibinfo{author}{Lin, G.}, \bibinfo{author}{Loy, C.C.}, \bibinfo{author}{Long, C.}, \bibinfo{author}{Li, Z.}, \bibinfo{author}{Zhao, R.}, \bibinfo{year}{2024}.
\newblock \bibinfo{title}{Eliminating feature ambiguity for few-shot segmentation}, in: \bibinfo{booktitle}{Proceedings of the European Conference on Computer Vision (ECCV)}, pp. \bibinfo{pages}{416--433}.
%Type = Inproceedings
\bibitem[{Xu et~al.(2023)Xu, Zhao, Lin and Long}]{Xu_2023_ICCV}
\bibinfo{author}{Xu, Q.}, \bibinfo{author}{Zhao, W.}, \bibinfo{author}{Lin, G.}, \bibinfo{author}{Long, C.}, \bibinfo{year}{2023}.
\newblock \bibinfo{title}{Self-calibrated cross attention network for few-shot segmentation}, in: \bibinfo{booktitle}{Proceedings of the IEEE/CVF International Conference on Computer Vision (ICCV)}, pp. \bibinfo{pages}{655--665}.
%Type = Article
\bibitem[{Yang et~al.(2020)Yang, Liu, Li, Jiao and Ye}]{pmm}
\bibinfo{author}{Yang, B.}, \bibinfo{author}{Liu, C.}, \bibinfo{author}{Li, B.}, \bibinfo{author}{Jiao, J.}, \bibinfo{author}{Ye, Q.}, \bibinfo{year}{2020}.
\newblock \bibinfo{title}{Prototype mixture models for few-shot semantic segmentation}.
\newblock \bibinfo{journal}{arXiv preprint arXiv:2008.03898} .
%Type = Inproceedings
\bibitem[{Yang et~al.(2023)Yang, Chen, Feng and Huang}]{Yang_2023_CVPR}
\bibinfo{author}{Yang, Y.}, \bibinfo{author}{Chen, Q.}, \bibinfo{author}{Feng, Y.}, \bibinfo{author}{Huang, T.}, \bibinfo{year}{2023}.
\newblock \bibinfo{title}{{MIANet}: Aggregating unbiased instance and general information for few-shot semantic segmentation}, in: \bibinfo{booktitle}{Proceedings of the IEEE/CVF Conference on Computer Vision and Pattern Recognition (CVPR)}, pp. \bibinfo{pages}{7131--7140}.
%Type = Inproceedings
\bibitem[{Zhang et~al.(2021)Zhang, Xiao and Qin}]{scl}
\bibinfo{author}{Zhang, B.}, \bibinfo{author}{Xiao, J.}, \bibinfo{author}{Qin, T.}, \bibinfo{year}{2021}.
\newblock \bibinfo{title}{Self-guided and cross-guided learning for few-shot segmentation}, in: \bibinfo{booktitle}{Proceedings of the IEEE/CVF Conference on Computer Vision and Pattern Recognition}, pp. \bibinfo{pages}{6933--6942}.
%Type = Inproceedings
\bibitem[{Zhang et~al.(2019)Zhang, Lin, Liu, Yao and Shen}]{zhang2019canet}
\bibinfo{author}{Zhang, C.}, \bibinfo{author}{Lin, G.}, \bibinfo{author}{Liu, F.}, \bibinfo{author}{Yao, R.}, \bibinfo{author}{Shen, C.}, \bibinfo{year}{2019}.
\newblock \bibinfo{title}{Canet: Class-agnostic segmentation networks with iterative refinement and attentive few-shot learning}, in: \bibinfo{booktitle}{Proceedings of the IEEE Conference on Computer Vision and Pattern Recognition}, pp. \bibinfo{pages}{5217--5226}.
%Type = Article
\bibitem[{Zhang et~al.(2020)Zhang, Wei, Yang and Huang}]{zhang2018sg}
\bibinfo{author}{Zhang, X.}, \bibinfo{author}{Wei, Y.}, \bibinfo{author}{Yang, Y.}, \bibinfo{author}{Huang, T.S.}, \bibinfo{year}{2020}.
\newblock \bibinfo{title}{Sg-one: Similarity guidance network for one-shot semantic segmentation}.
\newblock \bibinfo{journal}{IEEE transactions on cybernetics} \bibinfo{volume}{50}, \bibinfo{pages}{3855--3865}.
%Type = Inproceedings
\bibitem[{Zhu et~al.(2024)Zhu, Chen, Ji, Ye and Liu}]{Zhu_2024_CVPR}
\bibinfo{author}{Zhu, L.}, \bibinfo{author}{Chen, T.}, \bibinfo{author}{Ji, D.}, \bibinfo{author}{Ye, J.}, \bibinfo{author}{Liu, J.}, \bibinfo{year}{2024}.
\newblock \bibinfo{title}{{LLaFS}: When large language models meet few-shot segmentation}, in: \bibinfo{booktitle}{Proceedings of the IEEE/CVF Conference on Computer Vision and Pattern Recognition (CVPR)}, pp. \bibinfo{pages}{3065--3075}.

\end{thebibliography}
